# Impact of color and mixing proportion of synthetic point clouds on semantic segmentation

Shaojie Zhou [a,b], Jia-Rui Lin [a,b,*], Peng Pan [a,b], Yuandong Pan [c], Ioannis Brilakis [c]

[a] Department of Civil Engineering, Tsinghua University, Beijing, China, 100084

[b] Key Laboratory of Digital Construction and Digital Twin, Ministry of Housing and Urban-Rural Development, Beijing, China, 100084

[c] Department of Engineering, University of Cambridge, 7a JJ Thomson Avenue, Cambridge, CB2 1PZ, United Kingdom

**Abstract**

Deep learning (DL)-based point cloud segmentation is essential for understanding built environment. Despite synthetic point clouds (SPC) having the potential to compensate for data shortage, how synthetic color and mixing proportion impact DL-based segmentation remains a long-standing question. Therefore, this paper addresses this question with extensive experiments by introducing: 1) method to generate SPC with real colors and uniform colors from BIM, and 2) enhanced benchmarks for better performance evaluation. Experiments on DL models including PointNet, PointNet++, and DGCNN show that model performance on SPC with real colors outperforms that on SPC with uniform colors by 8.2 % + on both OA and mIoU. Furthermore, a higher than 70 % mixing proportion of SPC usually leads to better performance. And SPC can replace real ones to train a DL model for detecting large and flat building elements. Overall, this paper unveils the performance-improving mechanism of SPC and brings new insights to boost SPC's value (for building large models for point clouds).

**Keywords: Synthetic point clouds; Semantic segmentation; BIM; Deep learning; Mixing training**

## 1 Introduction

Buildings are the most frequently encountered environment in our daily lives. However, the built environment is often very complex, making it still a challenge for computers to accurately understand. With the development of deep learning, semantic segmentation of point clouds enables computers to understand simple built environment and has been widely applied in the fields of building and construction, including scan-to-BIM (building information modeling), construction progress and quality monitoring [1][2][3]. As a high-precision representation of the real world, point clouds contain wealthy but unordered 3D information. In recent years, a large number of deep learning algorithms have emerged for point cloud semantic segmentation tasks and have been widely applied in the field of building and construction, such as PointNet [4], DGCNN [5].

However, high-quality datasets are essential for the training of deep learning models, yet the creation and annotation of 3D datasets are frequently laborious and time-intensive endeavors. High-quality point cloud datasets are very scarce. Obtaining accurate point cloud data is challenging, necessitating the use of specialized equipment such as depth cameras or laser scanners. The cost of purchasing these devices is considerable, and scanning large-scale scenes is also a time-consuming process. Moreover, point cloud data also faces substantial labeling costs. For instance, the Stanford 3D Indoor Spaces Dataset (S3DIS) [6] contains nearly 700 billion points, each labeled with one of 13 classes, which indicates a huge labeling effort. Furthermore, the annotation tasks of buildings or construction sites are more complex and typically require specialized knowledge. This can be inferred from the annotation of 2D images: Duan [7] et al. spent more than 1800 h processing, annotating, and reviewing 21,863 construction images. The annotation of point clouds is more extensive, time-consuming, and challenging, making it a tedious and error-prone task.

To address this problem, methods such as self-supervised learning [8], transfer learning [9], and few-shot learning [10] have been proposed. However, they all aim to achieve better performance with a smaller amount of data or annotation, which still does not solve the problem of data scarcity. Therefore, some researchers have utilized synthetic data to compensate for the lack of training data, especially in the field of computer vision [11]. Moreover, the widespread application of BIM technology [12] has brought a large amount of 3D digital models in the field of building and construction, which also provides significant potential for the synthetic point cloud data of the built environment.

Previous research [13][14][15] has shown that synthetic point clouds generated based on BIM can improve the performance of 3D semantic segmentation models. However, pure synthetic data cannot replace real data to achieve comparable training performance [13][14], even synthetic point clouds with the same size and spatial distribution as real ones cannot achieve. Thus, there are still intrinsic gaps between the synthetic and real point clouds, which mainly depends on the spatial distribution and color differences. In addition, since synthetic point clouds cannot completely replace real point clouds for model training, the different mixing proportions of the real and synthetic datasets could play an important role in the model performance. Currently, there is still no clear explanation for the impact principles of the color and mixing proportion of synthetic point clouds.

Therefore, in order to deeply explore the impact of color and mixing proportion on model performance, this study first proposes a method for generating synthetic point clouds based on BIM. On this basis, two types of synthetic point cloud datasets and S3DIS real dataset are combined by different schemes to train three common 3D semantic segmentation models, for exploring the impact of color and mixing proportion of synthetic data. Specially, an additional set of benchmark experiments is proposed and conducted, making the comparison result more comprehensive and reliable. The remaining parts of this paper are organized as follows. Section 2 presents a literature review on 3D semantic segmentation models, synthetic point cloud generation, and training methods. Section 3 provides a detailed description of our synthetic point cloud generation method and briefly introduces the semantic segmentation models chosen in this paper. Section 4 conducts the experiments of different colors and mixing proportions. Section 5 further discusses the results and provides suggestions for synthetic point cloud generation and model training. Finally, Section 6 summarizes the overall paper.

## 2 Related work

### 2.1 3D semantic segmentation based on deep learning

Point cloud, one of the most used 3D data, can represent complex 3D spatial information but do not contain contextual information. Accurately extracting semantic information from point clouds is still a challenge. As an important means for computers to understand the real environment, 3D semantic segmentation based on deep learning has been widely applied in various fields. Usually, semantic segmentation methods are divided into four categories [16]: Projection-based Methods, Discretization-based Methods, Hybrid Methods and Point-based Methods. The unordered nature, interaction among points, and invariance under transformations make it difficult to learn features directly from point clouds [4]. Therefore, early methods tended to convert point clouds into forms that are easier for computers to process, such as voxel grids and images. But these methods inevitably result in information loss [17] and redundant storage [18]. Until Qi et al. [4] proposed the PointNet model, which pioneered an end-to-end point-based semantic segmentation approach. PointNet utilizes point-wise Multi-Layer Perceptron (MLP) methods to extract features from unordered point clouds. Subsequently, more and more point-based semantic segmentation networks emerged. PointNet++ [19] is a hierarchical network based on recursive PointNet, which further captures the local feature of point clouds. Based on random sampling and local feature aggregation modules, RandLa-Net [20] network efficiently learns features of complex point clouds. DGCNN[5] is a classical graph-based network that uses the edge convolution (Edge Conv) module to obtain local and global features of point clouds. MinkowskiNet [21] uses 4-dimensional convolutional neural networks for spatio-temporal perception, and has achieved excellent performance in indoor scene understanding. PCT [22], PointTransformers v1–3 [23][24][25] uses the transformer method to process point cloud sequences, achieving advanced performance in various 3D visual tasks. Here we only list some widely used semantic segmentation models, while other point-based methods also show strong potential in extracting point cloud features.

In terms of point cloud datasets, the datasets used for semantic segmentation are mainly outdoor scenes and urban scenes related to autonomous driving research. For the built environment, high-quality datasets such as S3DIS [6], ScanNet [26] are still scarce and singular, making it difficult to meet the diverse research needs of different fields.

### 2.2 Overview of synthetic point clouds

Due to properties of infinite generation and automatic annotation, synthetic data have been widely used in various computer vision tasks, such as object detection [27], pose estimation [28], semantic segmentation [29], and crowd counting [30]. In recent years, with the increasing accessibility of BIM and 3D models, more and more research focus on the synthetic point clouds and tries to utilize them to improve the training process of semantic segmentation deep learning models. Overall, the relevant research all focuses on two major aspects: the generation methods and effectiveness studies of synthetic data. The first is how to generate more realistic point clouds. And the second is the impact of synthetic data on the performance of deep learning models. The following section provides a detailed introduction to these two aspects.

#### 2.2.1 Generation of synthetic point clouds

Major synthetic point clouds are sampled from existing digital 3D models, commonly including online housing design models [9], randomly generated 3D models [31], and BIM models [13]. The spatial information of 3D models is similar to the real scene, providing raw materials for the creation of synthetic data. However, the spatial distribution of point clouds is random, and colors are easily influenced by the environment, making the realism of synthetic point clouds a major challenge. Overall, the framework of synthetic point cloud generation generally consists of three stages: (1) Geometric Generation, (2) Semantic Annotation, and (3) Color Assignment. Different studies have adopted different methods to achieve the above three steps. The first two steps are necessary, while the color assignment is optional. The following will provide a detailed introduction to these three steps.

**(1) Geometric Generation.** Methods in this step mainly include sampling and scanner simulation. Ma et al. (2020) [13] proposed a method to generate point clouds by uniformly sampling all objects of BIM models using FME WorkBench. This approach results in the points inside the objects are incorrectly sampled, which is significantly different from the real data. To overcome this problem, Zhai et al. [32] extracted surface geometric information by retrieving semantic properties and spatial topological relationships from IFC models. But this still not considered the spatial occlusion relationships between objects while real scanning. Tang et al. [14] proposed the Improved Hidden Point Removal (IHPR) algorithm to remove invisible points station-by-station after surface sampling. They all obtained synthetic point clouds with similar geometric features to the real scanning data. Furthermore, Ma et al. (2021) [33] abandoned the surface sampling method and instead used Blensor to simulate the manual scanning process by placing virtual scanners in the BIM space. Although the occlusion relationship is considered, the virtual scanner locations had to manually specified in this work, which was time-consuming when dealing with a large number of BIM models. Similarly, Noichl et al. [15] and Cazorla et al. [31] both used laser scanner simulation to create synthetic point clouds and further solved the problem of manually designating scanner locations. The former used fixed scanner locations, while the latter automatically selected scanner locations based on rules. Overall, these methods can effectively restore the spatial and occlusion relationship of points, but cannot process the measurement range of scanner. In addition, Song et al. [9] employed a CycleGAN neural network to transfer the spatial distribution from real point clouds to the uniformly sampled synthetic point clouds, but did not consider the occlusion relationships. Zhang et al. [34] obtained RGB images and depth maps by randomly placing virtual cameras in BIM models, and then reconstructed the point clouds from the depth information.

**(2) Semantic Annotation.** Due to the rich semantic information embedded in 3D models, semantic annotation of synthetic data is relatively straightforward. Except for early research [13] that requires manual annotation, major research automatically extracted and annotated semantic information from 3D models when generating geometric information of the point clouds. In addition, other research used methods based on projection [33] and ray tracing [15] to accomplish the annotation task.

**(3) Color Assignment.** Color is a key factor affecting the realism of synthetic point clouds. The methods used in different studies vary greatly. In the research of Ma et al. [13] and Noichl et al. [15], there were no colors in synthetic point clouds. Tang et al. [14] used colors exported from texture information of BIM models, which resulted in distortion compared to the real colors. Zhai et al. [32] colored points with BIM material and real colors, respectively, where the real colors were directly derived from S3DIS data. Furthermore, the CycleGAN neural network of Song et al. [9] also transferred the style of colors from real to synthetic data, but there were inevitably errors in the details. Furthermore, Zhang et al. [34] obtained RGB images of 3D models by using virtual cameras and added colors while reconstructing the point clouds from the depth maps. But this method depends on the realism of the rendered images. Despite the great efforts that researchers have made, generating synthetic point clouds with real colors remains a difficult task, which could have an impact on the performance of semantic segmentation models. Therefore, this study conducts a set of comparative experiments to explore the impact of colors on model performance.

Overall, the above methods are summarized and compared in the Table 1. In geometric generation, the occlusion relationship and measurement range of real scanner are the major factor to affect the realism of point clouds. On this basis, generation method in this research is devoted to consider these two factors. In addition, automatic semantic annotation is the biggest advantage of synthetic data, which is achieved in almost all research. As for colors of points, most research is no colors or single type of colors, insufficient to investigate the impact of color on model performance. However, this factor is the focus in this research.

**Table 1**

Comparison of synthetic point cloud generating methods

| Author | Data source | Geometric generation | Semantic annotation | Color assignment | |
|---|---|---|---|---|---|
| | | | | BIM color | Real color |
| Ma [13] | BIM model | | | | |
| Ma [33] | BIM model | ○ | √ | | |
| Song [9] | Online model | | √ | | √ |
| Cazorla [31] | Random model | ○ | | | √ |
| Zhai [32] | BIM model | | √ | √ | √ |
| Tang [14] | BIM model | ● | √ | √ | |
| Zhang [34] | BIM model | ● | √ | | √ |
| Noichl [15] | BIM model | ○ | √ | | |
| **Ours** | **BIM model** | ● | **√** | **√** | **√** |

* In geometric generation, ○ represents the occlusion is considered; ● represents the occlusion and measurement range are considered.

### 2.2.2 Synthetic point clouds for model training

After generating a large number of synthetic point clouds, the impact of synthetic data on model performance is the next step of research. In theory, the optimal scenario would be that synthetic data can entirely replace the real data and achieve performance no less than the real dataset. However, so far this is still an unrealistic expectation. There is an example exception. The synthetic dataset of Zhai et al. [32] was generated by BIM models identical to the S3DIS dataset and assigned colors the same as the S3DIS dataset, which ultimately achieved comparable training performance to the S3DIS dataset. Nevertheless, given that the synthetic dataset is too similar to the real dataset and unable to be obtained in practical application, the results are unfair and biased.

Consequently, most of the research prefer to mix synthetic datasets with real datasets for enhancing the training of segmentation models. Overall, there are two types of research strategies. The first method is to add synthetic data into a real dataset. For example, Tang et al. [14] added an equal amount of synthetic data to the real dataset. And the mixed training set demonstrated superior performance compared to the training set before mixing, proving the effectiveness of synthetic point clouds. The findings of Ma et al. [13] also showed that the performance was improved when real data are augmented with 1×, 2×, and 3× synthetic data, but none of them could match the performance of the real dataset with an equivalent total amount of mixed dataset. Nevertheless, the addition of synthetic data increases the size of the training set. So, the second strategy maintains the size of training set and changes the mixing proportions of two types of datasets. Noichl et al. [15] found that mixing 12 % of real data with 88 % of synthetic data resulted in slightly inferior performance than 100 % of real data. However, the optimal mixing proportion remains unclear. In the study by Zhai et al. [32], it was demonstrated that replacing a small amount of real data with synthetic data could achieve superior performance. Furthermore, Zhang et al. [34] mixed the S3DIS dataset with synthetic data at proportions of 0 %, 25 %, 50 %, 75 %, and 100 %, and ultimately found that the performance was positively correlated with the proportion of real data.

In summary, there is an inherent discrepancy between synthetic and real point clouds. Consequently, the mixed training of synthetic and real data is a more critical issue. However, previous studies didn't provide a comprehensive discussion on the mixing proportions. And nor did they implement benchmark experiments without synthetic data, thus failing to validate the value of synthetic point clouds. In light of the aforementioned research gap, this study conducts a series of experiments with a more comprehensive range of mixing proportions and simultaneously carries out additional benchmark experiments, to obtain more reliable results. Finally, based on these findings, an in-depth discussion is conducted to offer suggestions for research on synthetic point clouds.

## 3 Methodology

In order to investigate the impact of synthetic point cloud color (SPC color) and mixing proportion with real data (SPC proportion) on the performance of semantic segmentation models. This study first proposes a synthetic point cloud generation method based on the BIM model. And on this basis, experiments are conducted on different training sets of synthetic and real dataset. Fig. 1 shows the methodology of our research. It contains two key parts. The first part is to generate synthetic point clouds from BIM models. And in this study, two types of synthetic datasets are generated with different colors. Then the second part involves two sets of experiments. The variable in Experiment 1 is SPC color and in Experiment 2 is SPC proportion, for exploring their impact on model performance. This section provides a detailed description of the methodologies for synthetic point cloud generation, the semantic segmentation models, and the experimental design.

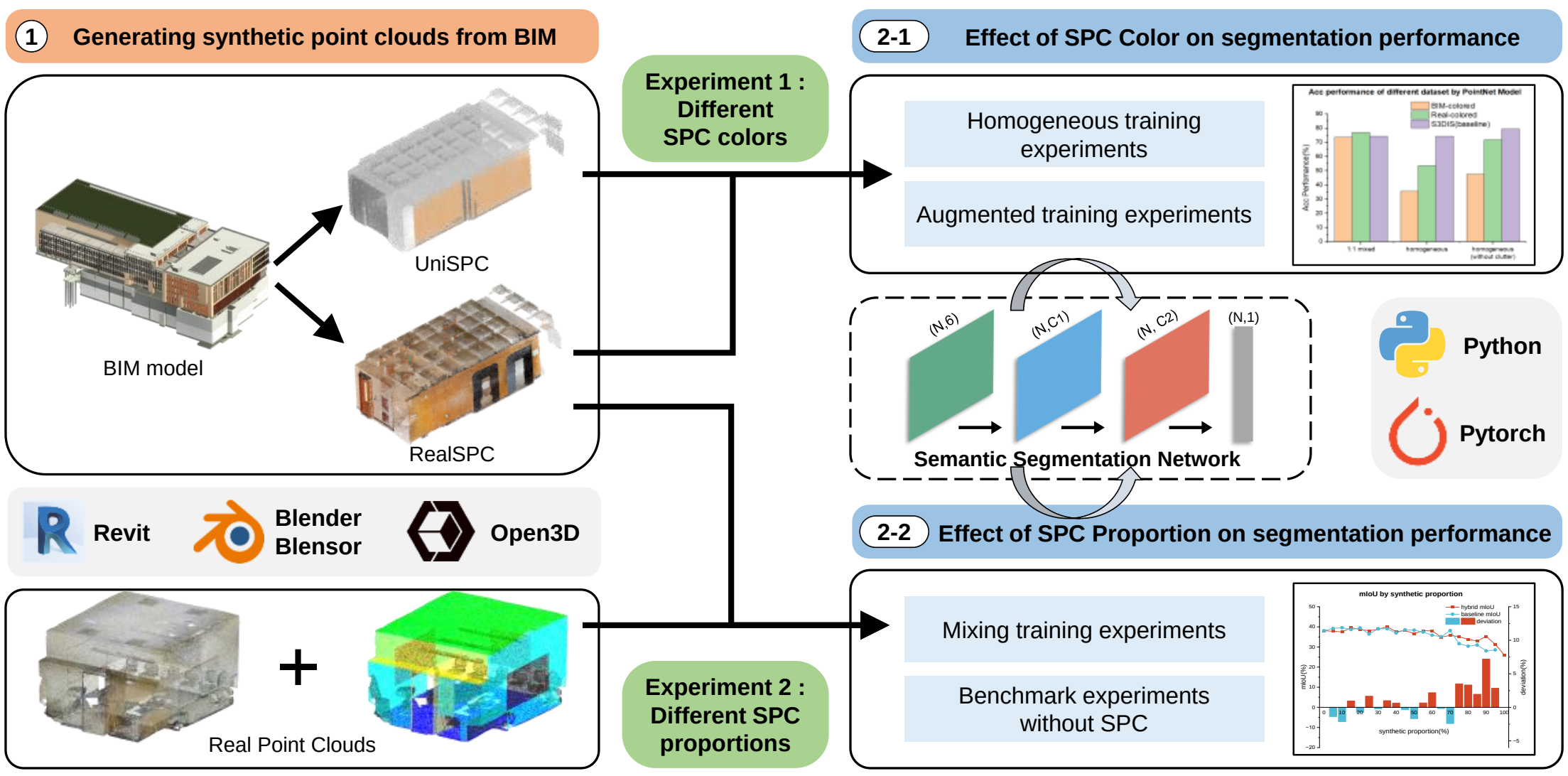

* SPC is an abbreviation for synthetic point cloud(s).

**Fig. 1.** Methodology. 1) Synthetic point clouds generation, and 2) Experiments on effect of color and mixing proportion.

### 3.1 Synthetic point cloud generation

Synthetic point cloud generation is the foundation of the entire research. In this study, a method based on scanner simulation has been proposed. The biggest innovation of the method is to automatically simulate the scanner layout of real scanning, both considering the occlusion relationship and measurement range of real scanners. The whole framework is illustrated in Fig. 2.

**Step 1: Geometric Generation.** In order to simulate the actual scanning process: point clouds are composed of multiple scans and splices, the Industry Foundation Classes (IFC) is used to extract the planar contours of each scene in BIM models. Then within these contours, uniformly distributed scanner stations can be determined by Centroidal Voronoi Tessellation (CVT) algorithm. On this basis, synthetic point clouds containing XYZ information can be automatically generated by virtual scanners at each station in the Blensor.

**Step 2: Semantic Annotation.** In this step, the dense point clouds containing semantic information are exported from the BIM models and registered with the point clouds generated in Step 1. The overlapping parts are then annotated with the same semantic information to obtain the annotated synthetic point clouds.

**Step 3: Color Assignment.** In this work, two methods are employed to color synthetic point clouds. The first method is as same as Step 2. Given that BIM models contain both semantic and textural information, the colors can be assigned while synthetic data is annotated. However, this color originates from the "consistent color" in the BIM model, resulting in each component being unicolored. Consequently, there is a discrepancy between this color and the real texture. This point cloud is designated as an unicolored synthetic point cloud (denoted as UniSPC). The alternative method is to register the synthetic point clouds with the scanned point clouds and share the distribution of colors. This type of synthetic point cloud is colored with real colors and is designated as a real-colored synthetic point cloud (denoted as RealSPC).

The detailed methods for the three key steps are as follows.

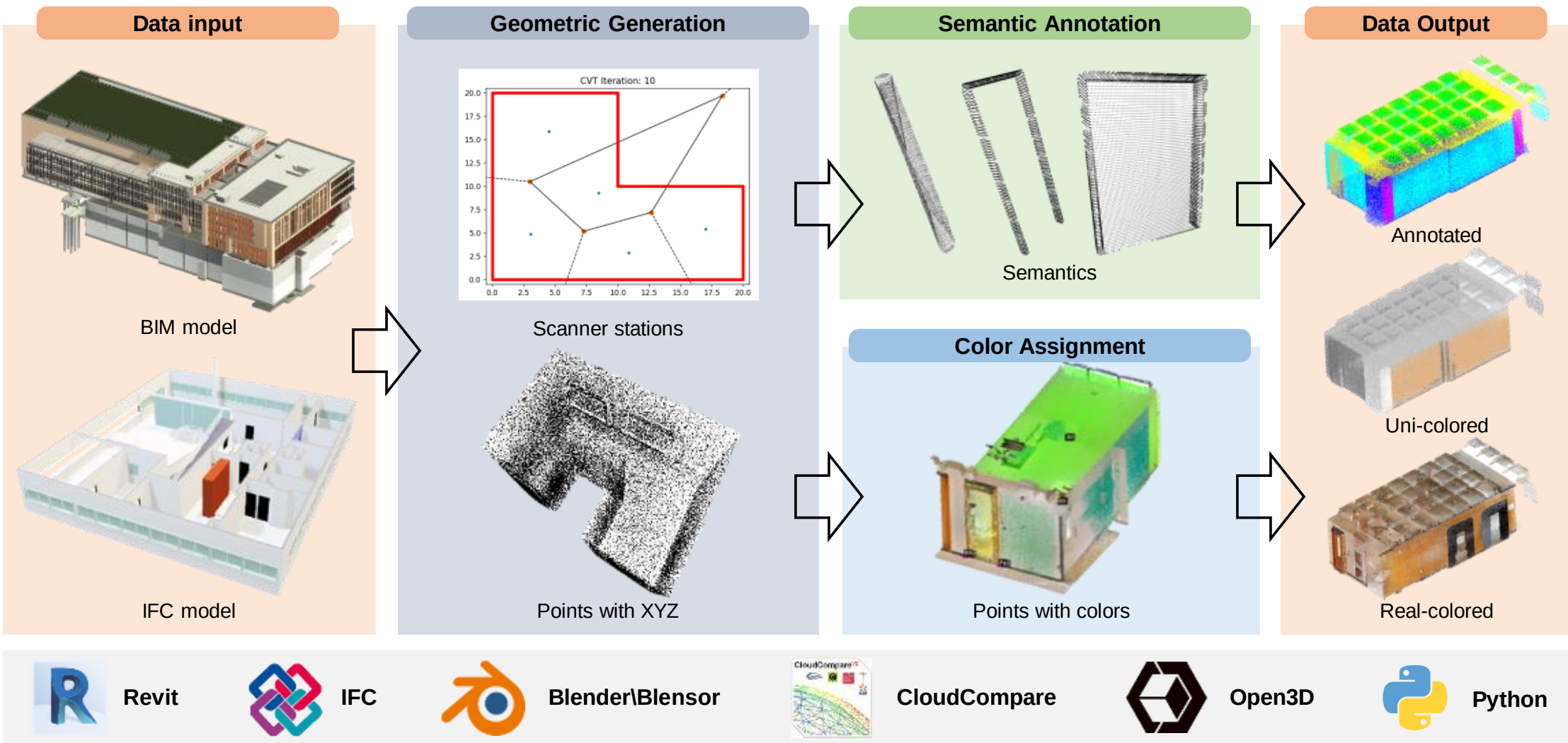

**Fig. 2.** Framework of synthetic point clouds generation.

### 3.1.1 Geometric generation

In this study, BIM modeling is carried out in Revit. As a kind of 3D model, the Revit model (uniformly denoted as BIM model in the following text) contains accurate spatial information, which can be used to generate synthetic point clouds. However, it should be noted that the real point cloud only represents the surface information of objects and involves complex occlusion relationships. Consequently, the BIM model cannot be converted into a point cloud without processing. To address this issue, a feasible approach is to utilize the scanner simulation method to extract the spatial information of BIM models. And Blensor (https://www.blensor.org/) is a free open-source simulation package that allows users to simulate the major ranging sensors. Given the simulation cost, Time-of-Fight (ToF) camera is chosen as the virtual sensor in this research.

Furthermore, the actual sensors always have a specific measuring range, making traditional scanned point clouds have to sample and register station-by-station. And the selection of scanner stations must consider factors such as the measuring range and the scale of the built environment, and theoretically need to evenly distribute within the space. In order to simulate this process, our data generation is also divided into a series of station-by-station scanning. So, it is necessary to identify the scene boundaries and scanner stations in BIM models before scanning. In reference to the scenes in S3DIS dataset, we select spaces such as office, conference room, and hallway as a single scene for scanning. This space information is specifically defined in the BIM model as a component named “Room”. As shown in Fig. 3 (a), the “Basic Sample Project” provided by Revit includes room components such as bath and hall, corresponding to the smallest unit of scanning. To extract this room information, BIM models are converted to the Ifc format, where room information is stored in the property IfcSpace.

IfcSpace represents an area or volume bounded actually or theoretically. As shown in Fig. 3 (b) and Fig. 3 (c), space represented by IfcSpace is extruded from a closed IfcPolyLine (footprint) on a plane. Therefore, the scene boundary in BIM model is as same as the footprint in IfcSpace. The latter one can be easily extract by a Python script based on IfcOpenshell package (http://ifcopenshell.org/).

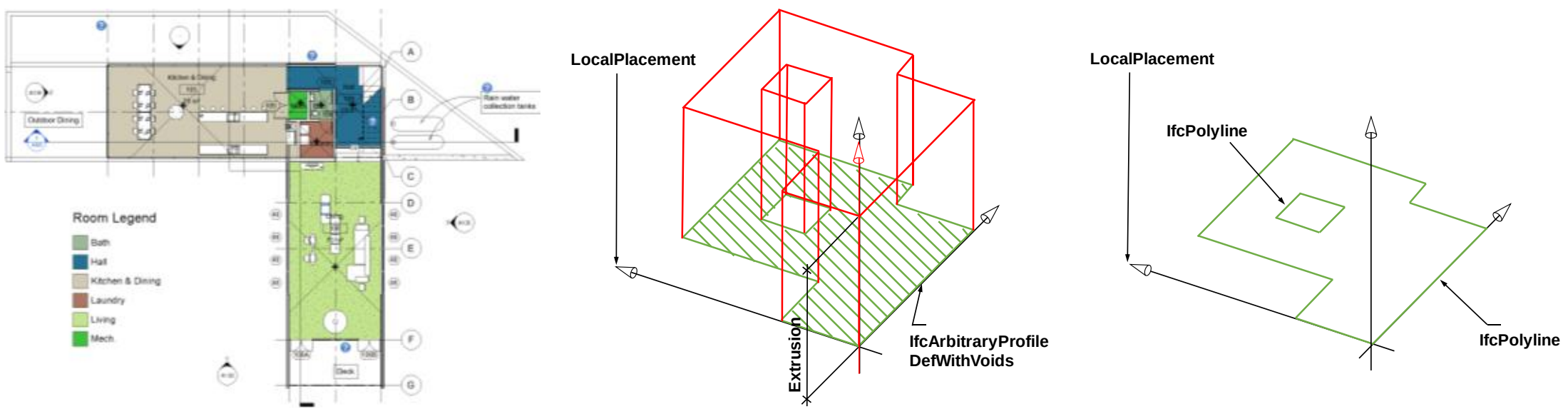

(a) (b) (c)

**Fig. 3.** Room component and IfcSpace. (a) Room components in “Basic Sample Project” of Revit. (b) IfcSpace is an extrusion of footprint into the swept area solid. (c) Space footprint of IfcSpace [35].

On this basis, the problem of determining scanner locations is converted to a math problem of uniform tessellation within a closed boundary. A feasible solution is centroidal Voronoi tessellation (CVT) in computer graphics. Voronoi tessellation can be described as a partition of a plane divided into regions ($V = \{V_1, V_2, \ldots, V_n\}$) close to each of a given set of seed points ($C = \{c_1, c_2, \ldots, c_n\}$), and satisfy the following equation:

$$V_i = \{x \in \mathbb{R}^2 \mid \|x - c_i\| \leq \|x - c_j\|, \forall\, j \neq i\}$$

CVT is a specific type of Voronoi tessellation in which the seed point of each Voronoi cell is also its centroid, making the regions divided from closed polygons being approximately the same size. The Lloyd's algorithm is a frequently employed algorithm for the generation of CVT, which is solved by iteratively calculating the centroid of the Voronoi diagram. Fig. 4 shows the process of CVT generation for two distinct scene boundaries based on the Lloyd algorithm. In this study, Lloyd's algorithm is used to automatically obtain a given number of scanner stations within the boundaries of IfcSpace, thereby preparing for the point cloud generation in Blensor. Specifically, in order to ensure that the synthetic point clouds generated from different scenes have similar density, the initial numbers of scanner stations and iterations in this algorithm are positively correlated with the area size enclosed by the scene boundaries.

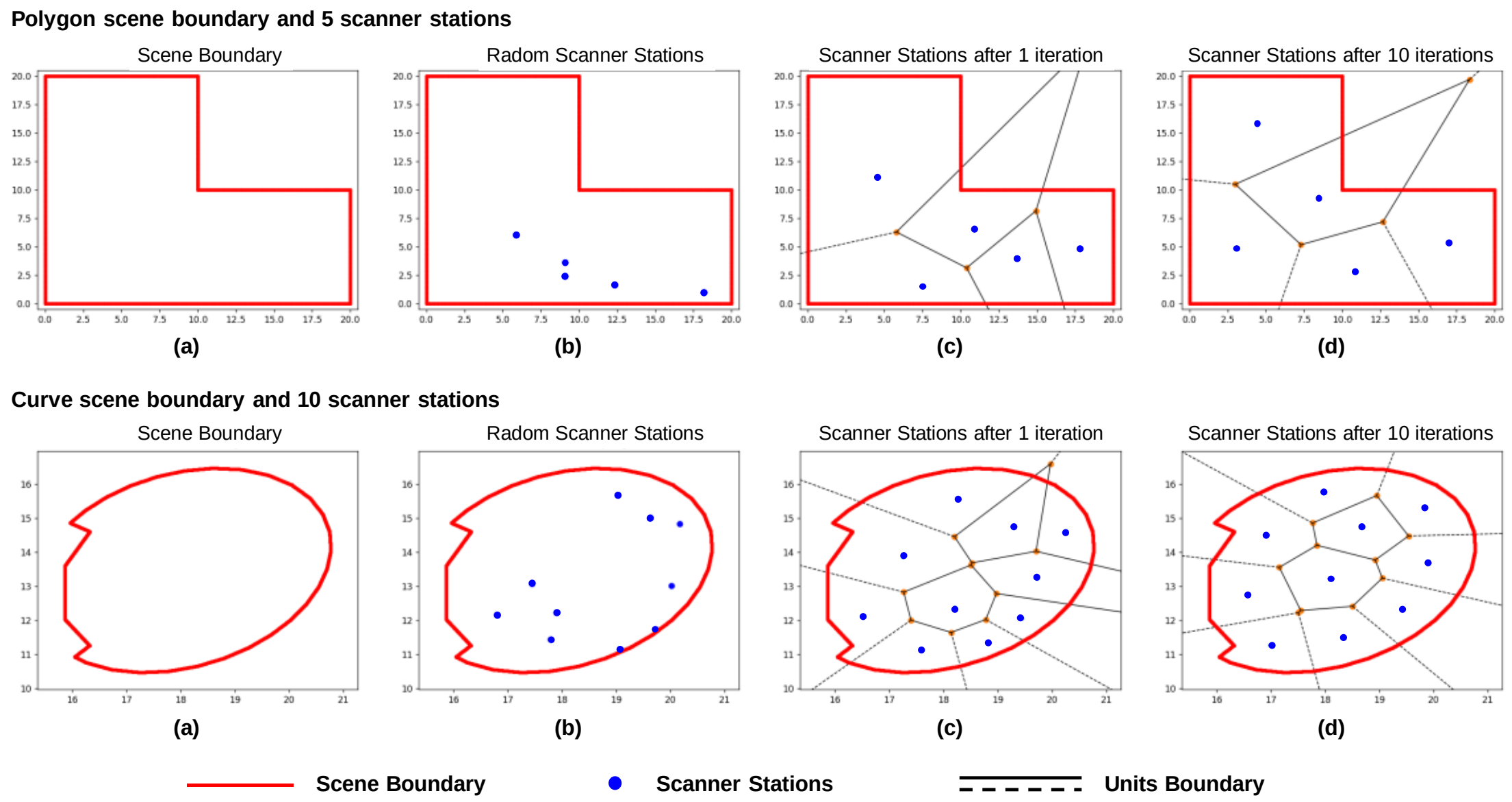

**Fig. 4.** Visualization of Lloyd's algorithm (a) Scene boundary. (b) Initial random scanner stations. (c) Algorithm iteration once. (d) Algorithm iteration 10 times.

Finally, the BIM models are imported into the Blensor. A Python script reads the coordinates of the scanner stations within each IfcSpace boundary, and sets the location of virtual sensors at the corresponding floor (height). Subsequently, the script automatically invokes the Blensor API to scan and register point clouds station-by-station, and finally saves them in the same format as S3DIS. There are no reference values for the parameters of the virtual ToF sensor. So, the optimal sensor parameters are configured with the objective of the ideal density of synthetic point clouds. The main parameters of the virtual sensor are presented in Table 2.

Based on the above method, synthetic point clouds can be automatically generated from BIM models. These synthetic data reflect the true geometric coordinates of real scene, but no color or semantic information. And this is addressed in the subsequent two stages.

**Table 2**

Main parameters of virtual scanner in Blensor

| **Parameter** | **Value** | **Description** |
| --- | --- | --- |
| SCAN_STEP_W | 25° | Horizontal scanning angle step size |
| SCAN_STEP_H | 25° | Vertical scanning angle step size |
| SCAN_NUM_W | 15 | Number of horizontal scans |
| SCAN_NUM_H | 6 | Number of vertical scans |
| TOF_XRES | 20pixel | Horizontal resolution |
| TOF_YRES | 20pixel | Vertical resolution |
| TOF_LENS_ANGLE_W | 30° | Horizontal field of view angle |
| TOF_LENS_ANGLE_H | 30° | Vertical field of view angle |
| TOF_MAX_DIST | 6.5 m | Maximum measurement distance |
| TOF_FOCAL_LENGTH | 2.0 m | Focal length |

### 3.1.2 Semantic annotation

The rich semantic information contained in BIM models can comprehensively represent elements of the built environment. Therefore, this study proposes an automatic annotation method for synthetic data based on BIM models. The core process is divided into two steps.

Step 1: Semantic extraction. Firstly, the BIM model is imported into Blender. Then based on the Open3D [36] plugin, a Python script traverses through all components in the BIM model and exports each component separately as a dense point cloud. Of course, the exported files are named with category labels, to ensure that the synthetic data can be properly labeled.

Step 2: Overlapping annotation. Since the synthetic point clouds (with XYZ coordinates) and the exported component point clouds belong to the same coordinate space, their overlapping parts share the same semantics. Assuming the synthetic point cloud is the source point cloud $P_s$ and the exported component point cloud is the target point cloud $P_t$, the distance for each point $p_i \in P_s$ to the closest point $p_j \in P_t$ can be used to judge the overlapping degree of them. Considering specific $p_i \in P_s$ and arbitrary point $p_j \in P_t$, the $d_i$ can be calculated by the following equation.

$$d_i = \min_{p_j \in P_t} \left\| p_i - p_j \right\|_2$$

And the corresponding index $j^*$ can be represented as:

$$j^* = \arg\min_{p_j \in P_t} \left\| p_i - p_j \right\|_2$$

If the distance is shorter than threshold (represented as $d_i < t$), point $p_i$ and point $p_{j^*}$ are considered to have the same spatial location. Then the label $C_{p_{j^*}}$ of point $p_{j^*}$ (BIM point) are pass to the label $C_{p_i}$ of point $p_i$ (synthetic point). Otherwise, the label $C_{p_{j^*}}$ is set as clutter.

$$C_{p_i} = \begin{cases} C_{p_{j^*}}, & d_i < t \\ clutter, & d_i \geq t \end{cases}$$

The above calculation process is integrated into a Python script. By traversing both synthetic and component point clouds, each point in synthetic data is annotated with a category label.

### 3.1.3 Color assignment

Color is crucial and the most challenging information in synthetic point cloud generation. Notably, BIM models contain not only semantic information but also simple textured colors. Therefore, the semantic annotation method introduced in Section 3.1.2 can also be used to color the synthetic point cloud. However, these colors are the consistent colors of BIM models, where a single component is unicolored with a distorted color. Compared to the real texture, consistent colors are very different from scanned colors. For convenience, this simple and unicolored point cloud with consistent colors of BIM models is referred to as Unicolored synthetic point clouds (UniSPC).

In order to generate synthetic point clouds with a more realistic color, this study also utilizes scanned point clouds to color the synthetic ones. The whole process is as follows: (1) select synthetic point clouds and real scanned point clouds of the same scene; (2) use the CloudCompare software to manually register the two sets of point clouds; (3) share the colors of the overlapping parts. This part is similar to the Section 3.1.2. In this part, the threshold must be selected anew. And the RGB values of colors are passed instead of the category labels. However, the drawbacks of this method are also obvious, as both BIM models and scanned point clouds of the same scene are required. Since hardware equipment has become more accessible today, although this is still a laborious task, but no longer a difficult task. This method provides an effective solution for creating high-quality synthetic point clouds, which can be used for synthetic color research. In our study, this type of point clouds with real colors is denoted as Real-colored synthetic point clouds (RealSPC). And the above two types of datasets are the main research objects in subsequent experiments.

## 3.2 Semantic segmentation model

This study selects three typical point-based networks to evaluate the value of synthetic point clouds on semantic segmentation using deep learning, namely PointNet [4], PointNet++ [19], and DGCNN [5]. These three models are relatively simple and have high computational efficiency, so they are widely used in the field of building and construction [37][38][39][40]. The evaluation results on these models can further explore the engineering applications of synthetic point clouds. They all use raw point cloud data as input and output the category label of each point to achieve the segmentation task. PointNet is the earliest deep learning network based on points. Furthermore, it employs point-wise MLP and max pooling to extract feature information from points, overcoming the unordered challenge of point cloud data. But in the PointNet, the features of points are independent of each other and lack the extraction of local features between adjacent point clouds. So, the PointNet++ network proposes a hierarchical feature learning framework, which can learn local features at different levels in the neighborhood of point clouds. In terms of the DGCNN network, the topological structure is constructed in the feature space of points and the local features are aggregated by EdgeConv modular. The biggest characteristic of DGCNN is the dynamically updated graph structure, which enables the network to extract semantic information in a deeper feature space. This method has been proven to be superior to feature aggregation in traditional geometric spaces, such as PointNet++. The above three networks are widely used and are chosen in the subsequent experiments to assess the value of synthetic data.

### 3.3 Experiment design

The objective of this study is to investigate the impact of color and mixing proportion of synthetic point clouds on the performance of segmentation models. To this end, two sets of experiments have been designed as follows.

**Experiment 1: effect of SPC Color on segmentation performance.** In this series of experiments, two types of training sets are employed to investigate the impact of color differences in synthetic point clouds: the homogeneous training set and the augmented training set. In the homogeneous training experiment, different types of synthetic datasets are utilized as a single training set to directly train the segmentation model, and compared with a real data training set of the same size. In the augmented training experiment, a small amount of real data is added to the synthetic data for augmentation, and the performance is compared with that of the non-augmented training set. This series of experiments aims to validate the effectiveness of synthetic point clouds and to analyze the impact of color on segmentation performance.

**Experiment 2: effect of SPC Proportion on segmentation performance.** In this set of experiments, real data is gradually replaced by synthetic data, to form training sets of different mixing proportions. In order to obtain a more comprehensive result, mixing proportions with an interval of 5 % are adopted. It is noteworthy that the mixing proportion in this study is defined as the proportion of synthetic point clouds to the total number of point clouds in the training set (proportion of scene quantity). Furthermore, to quantify the value of synthetic data in mixed training sets, a series of benchmark experiments is conducted, in which synthetic point clouds are removed from the mixed training set. Ultimately, for each model, 21 mixing experiments (with mixing proportions from 0 to 100 %) and 19 benchmark experiments (with corresponding mixing proportions from 5 % to 95 %) are carried out. And the remaining two benchmark experiments are not considered: the experiment with 0 % mixing proportion contains no synthetic data, and the experiment with 100 % mixing proportion entirely consists of synthetic data. The objective of this series of experiments is to investigate the impact of mixing proportion on segmentation performance.

It's worth noting that benchmark experiments without synthetic point clouds in Experiment 2 are essential. Given that the effect of the number of real data on model performance is not clear, it's conceivable that the changes in performance with different mixing proportions could be attributed to the changes in the number of real data, rather than the critical role played by synthetic point clouds. Experiments on mixing proportions lacking benchmark experiments may be unreliable. This is crucial but missing in previous studies. So, benchmark experiments in this study ensure greater credibility of the experiments and quantify the value of synthetic data.

In order to compare the effects of different training sets, Mean Intersection over Union (mIoU) and Overall Accuracy (OA) are used to evaluate the overall performance of the model. At the same time, Intersection over Union (IoU) and Accuracy (OA) are used to evaluate the performance of individual class. These metrics are commonly used for the task of point cloud segmentation and are widely used for evaluating synthetic point clouds datasets [14][15][32][33].

## 4 Experiments and results

### 4.1 Data preparation

Before experiments, four synthetic point cloud datasets are prepared in this study. Two of them are generated by our method in Section 3.1, while the remaining two are sourced from Tang's study [14]. In addition, given that S3DIS is the most used indoor dataset and has a large scale, it's utilized to validate the effectiveness of the synthetic data and investigate the impact of SPC proportion on segmentation performance. S3DIS dataset is commonly used in studies on segmentation of built environment. It contains 6 areas with 271 scenes. Each point is annotated with one of the 13 semantic classes (wall, column, beam, ceiling, floor, door, window, sofa, board, chair, desk, bookshelf, and clutter). The summary of the above five datasets is presented in Table 3.

**Table 3**

Number of classes and scenes in five datasets

| Type | Source | Name | Classes | Scenes |
|---|---|---|---|---|
| Real dataset | manual scanning | S3DIS | 13 | 271 |
| Synthetic dataset (Tang et al.) | BIM model same with S3DIS | IS-PtC | 13 | 44 |
| | | FS-PtC | 13 | 44 |
| Synthetic dataset (**Ours**) | BIM model of NCEB | UniSPC | 8 | 256 |
| | | RealSPC | 8 | 44 |

In this study, 256 scenes of synthetic point clouds are generated from the BIM model of a newly-built office building (NCEB in Tsinghua University). So, the scenes of these data are obvious different from the scenes in S3DIS, which is a characteristic of our synthetic dataset. These synthetic data is colored by the consistent colors from the BIM model, so called this dataset UniSPC. Furthermore, in order to generate point clouds with more realistic colors, we manually scanned the fourth floor of the building and colored partial synthetic data based on these real data. Although this method greatly reduces the difficulty of obtaining the dataset, it still requires time and labor costs. From a research perspective, finally, 44 scenes of synthetic point clouds with real colors are generated, and this dataset is referred to as RealSPC. The scenes of these synthetic point clouds include offices, conference rooms, washrooms, hallways, etc., relying on the division of "room components" in BIM models. The annotation labels also refer to S3DIS. However, furniture, such as sofas, boards, chairs, desks, and bookshelves, are not included in the BIM models. Therefore, in our synthetic data, points are only divided into 8 classes: ceiling, floor, wall, beam, column, window, door, and clutter, not including other furniture. Visualization of partial scanned point clouds, UniSPC, RealSPC and annotated point clouds is shown in Fig. 5.

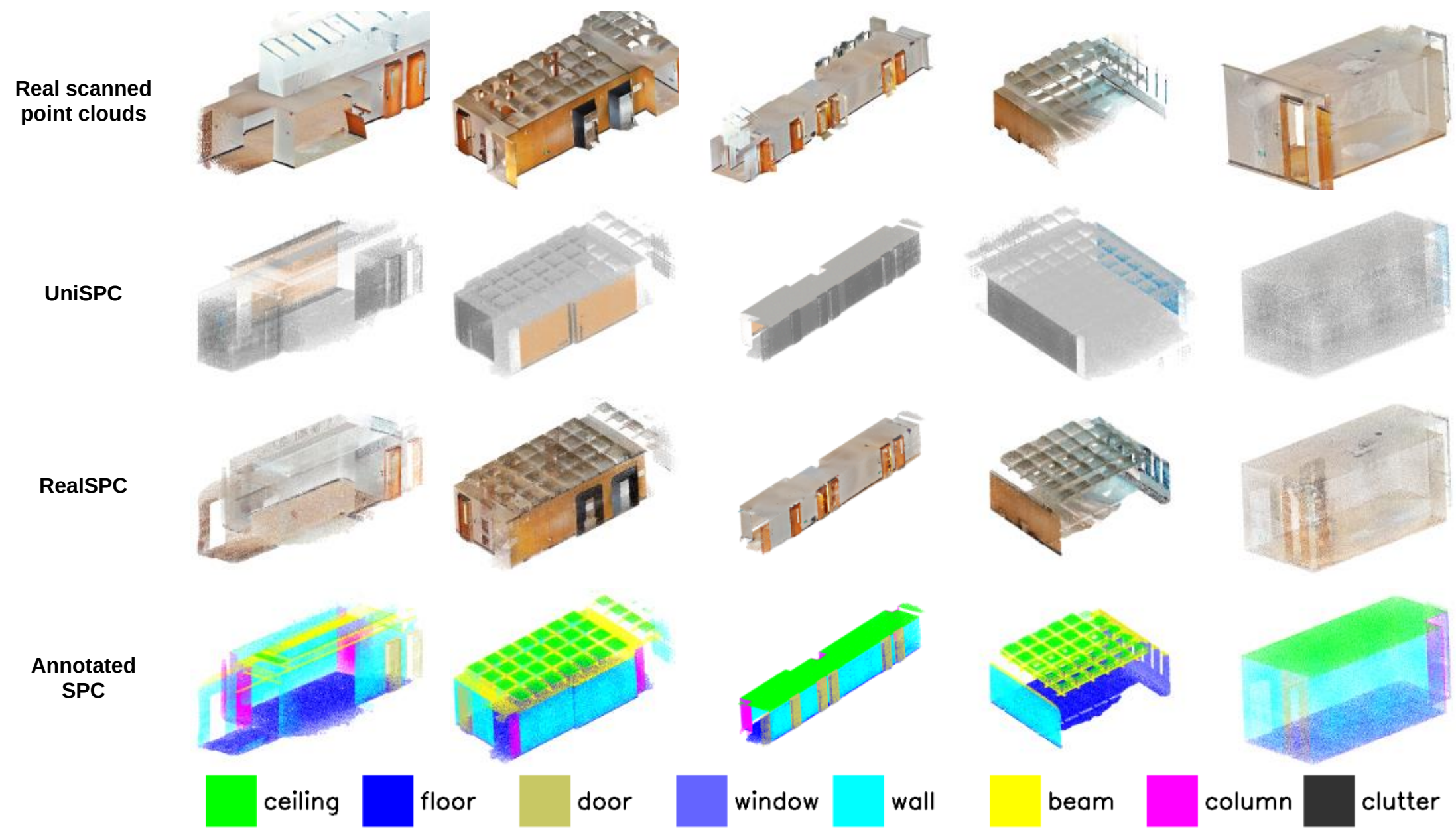

**Fig. 5.** Visualization examples of scanned point clouds, UniSPC, RealSPC and annotated point clouds.

Additionally, to validate the effectiveness of our synthetic point clouds, the BIMSyn dataset (an open-sourced synthetic point cloud dataset proposed by Tang et al. [14]) is also incorporated into the experiments of this study. BIMSyn is likewise a synthetic data generated based on BIM. However, the spatial distribution of its BIM models is the same as S3DIS (44 scenes in Area 1). Each point is also labeled as one of the 13 categories. BIMSyn contains two types of synthetic point clouds: IHPR-based synthetic point cloud (IS-PtC) and fully synthetic point cloud (FS-PtC). The former is generated by the IHPR algorithm proposed by the authors, considering the occlusion relationship between objects, and removing invisible points. The latter, however, is a complete point cloud of the entire scene without removing any points, thus having more complete local spatial features. In the study by Tang, the mixed datasets of these two types of synthetic data with S3DIS were demonstrated to enhance the segmentation performance of three different models. So, these two synthetic datasets are considered effective on model training and will also be used for model training in the following experiments.

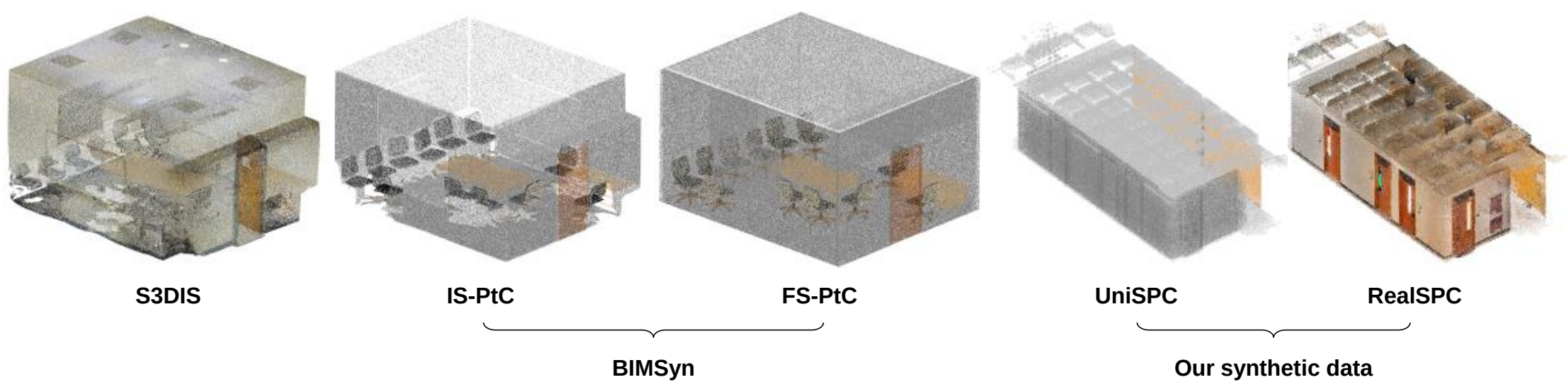

**Fig. 6.** Visualization of five types of datasets

In summary, a total of 5 datasets have been prepared for our experiments. And the visualization of them is shown in Fig. 6. Notably, our synthetic point clouds only contain 8 distinct labels and don't include various types of furniture. Therefore, in order to ensure that the segmentation performance is comparable, the annotation of both S3DIS dataset and BIMSyn datasets is aligned to 8 classes by maintaining 7 classes unchanged (ceiling, floor, wall, beam, column, window, and door) and consolidating the remaining classes into the clutter. As shown in Table 4, in all datasets, the proportions of the main building components such as ceilings, floors, and walls are relatively high, while the proportions of other components are relatively low. However, since the real scene is filled with furniture, the proportion of clutter in the real dataset is relatively high. While there is almost no clutter in our synthetic dataset.

**Table 4**

Proportion of points in different datasets (%)

| Dataset | ceiling | floor | door | window | wall | beam | column | clutter |
|---|---|---|---|---|---|---|---|---|
| S3DIS | 19.3 | 16.5 | 4.8 | 2.5 | 27.8 | 1.7 | 2.0 | 25.4 |
| IS-PtC | 19.7 | 15.5 | 6.3 | 0.5 | 34.3 | 4.8 | 3.2 | 15.7 |
| FS-PtC | 14.2 | 14.6 | 4.3 | 2.4 | 37.4 | 3.9 | 3.6 | 19.6 |
| UniSPC | 19.6 | 21.0 | 3.3 | 6.1 | 35.3 | 11.0 | 3.7 | 0.1 |
| RealSPC | 19.4 | 23.7 | 4.9 | 4.3 | 32.0 | 11.7 | 4.0 | <0.1 |

### 4.2 Model configuration and performance evaluation

In this study, synthetic data and real data are combined to form different mixed datasets for the training of PointNet, PointNet++ and DGCNN models. Before implementing the experiments, the following agreements are reached.

**Model configuration:** During the training phase, only the number of units in the output layers of three networks is modified from 13 to 8. While the remaining network architecture and training parameters are maintained at their default settings. Given that the point clouds are down sampled into blocks of 1m*1m for both training and testing in three networks, a fixed random seed is employed for each network during the experiments. This ensures that the data input is consistent across training and testing phases, thereby minimizing the interference of random factors.

**Performance evaluation:** Each type of model is trained for the same number of epochs, with PointNet and PointNet++ being trained for 32 epochs and DGCNN for 50 epochs. Taking the 5 % mixing proportion as an example, as shown in Fig. A, the training of all models tends to be sufficient. The outcome of each training epoch is validated on a dataset containing 40 scenes of real point clouds from Area 2 of S3DIS. Finally, the segmentation performance is defined as the evaluation metrics of the epoch with the highest mIoU on the validation set. Notably, given that our synthetic point clouds lack clutter class, so in addition to the traditional evaluation metrics OA and mIoU, an accuracy-relevant metric (denoted as $OA_7$) is proposed to assess the contribution of our synthetic data to the segmentation performance. $OA_7$ represents the proportion of correct predictions (among 7 classes of points without clutter) in all 7 classes of points without clutter. The calculation method of this metric is identical to OA, just not considering the clutter class. The OA and $OA_7$ are calculated as follows, respectively:

$$OA = \sum_{i=1}^{8} p_i \cdot A_i \, / \sum_{i=1}^{8} p_i = \sum_{i=1}^{8} p_i \cdot A_i$$

$$OA_7 = \sum_{i=1}^{7} p_i \cdot A_i \, / \sum_{i=1}^{7} p_i$$

where $p_i$ represents the proportion of points of $i^{th}$ class, and $A_i$ represents the corresponding prediction accuracy. Specially, $i = 8$ corresponds to the clutter class. From the formula above, it can be seen that OA and $OA_7$ metrics have the same preference in dealing with “imbalanced data”. So, $OA_7$ metric is reasonable for presenting the model prediction ability of the remaining 7 classes. Given that there are almost exclusively these seven classes of points, $OA_7$ can objectively evaluate the ability of our synthetic dataset.

## 4.3 Experiment 1: Effect of SPC color on segmentation performance

### 4.3.1 Homogeneous training experiment

To validate the effectiveness of our synthetic dataset and to investigate the impact of SPC color on the segmentation performance, a series of experiments using homogeneous training sets is first conducted. Our UniSPC and RealSPC, FS-PtC and IS-PtC from BIMSyn, as well as S3DIS, five different types of point clouds are used as single training sets to directly train three models. Each dataset comprises 44 scenes of point clouds. In the real datasets, all data originates from Area 1 of the S3DIS. In the UniSPC, 44 scenes of point clouds are randomly selected from 256 scenes. Finally, the training results of five experiments are all evaluated on the test set of 40 scenes of real point clouds from S3DIS Area 2.

The evaluation metrics of the above experiments are calculated and shown in Table 5. In terms of overall performance, none of the synthetic datasets can achieve performance comparable to those of S3DIS. This outcome is not unexpected and indicates that there is still a significant gap in features between synthetic and real point clouds. When comparing the four types of synthetic datasets, as shown in the left of Fig. 7, it is evident that the RealSPC demonstrates superior performance across all three networks, with an increase of 4.2 %, 11.8 %, 5.9 % in mIoU and 4.8 %, 21.2 %, 16.4 % in OA over the other three synthetic datasets with colors from BIM models, respectively. Compared to UniSPC, RealSPC has improved 8.2 % + in mIoU and 18.0 % + in OA, respectively. And the UniSPC achieves comparable results with two synthetic datasets from BIMSyn. The findings confirm the effectiveness of our synthetic point clouds and reveal the significant role of color in segmentation model training. Despite entirely different geometric generation methods and spatial distribution of them, three synthetic point clouds with consistent colors of BIM models have comparable effects. However, the RealSPC has shown the best performance in mIoU, $OA_7$ and OA metrics all, indicating that segmentation networks are sensitive to the color features, and more realistic colors can significantly enhance the model performance.

Furthermore, in theory, the capability of three networks to extract features is progressively enhanced, causing the model performance improving progressively. The performance of S3DIS and RealSPC both substantiate this opinion. However, similar conclusions cannot be drawn from the remaining three synthetic datasets with consistent colors of BIM, and even the performance of UniSPC shows an opposite trend. This finding also illustrates that color plays a crucial role in both the local and global features of point clouds. Incorrect colors can lead the model to aggregate wrong features, which in turn results in a decline in the segmentation performance.

Notably, if we use $OA_7$ to evaluate the accuracy for the seven classes without clutter, as shown in the right of Fig. 7, our RealSPC achieves performance close to S3DIS training set, with gaps of 5.4 %, 0.5 %, and 2.3 %, respectively. This suggests that these seven class points of our RealSPC exhibit features similar to those of real point clouds. This finding confirms the effectiveness of our synthetic data and generation method again, which successfully simulates the real scanned point clouds. However, for the same metric $OA_7$, UniSPC with consistent colors has poor performance, highlighting the significant role of color in model training. Furthermore, this result also indicates the generalization capability of models for segmenting different scenes with similar components. The seven types of points are all common architectural components in built environment. Although the BIM model used in this study is a building in our university, which differs from the real buildings of S3DIS, the synthetic point clouds finally still demonstrate a similar ability in predicting these seven classes.

**Table 5**

Model performance on five homogenous training sets (%)

| Source | Training set | PointNet | | | PointNet++ | | | DGCNN | | |
|---|---|---|---|---|---|---|---|---|---|---|
| | | **mIoU** | **OA** | **$OA_7$** | **mIoU** | **OA** | **$OA_7$** | **mIoU** | **OA** | **$OA_7$** |
| Real | S3DIS | 38.1 | 73.5 | 77.2 | 43.3 | 77.4 | 77.4 | 44.3 | 79.5 | 83.2 |
| BIMSyn | FS-PtC | 14.7 | 31.3 | 24.3 | 10.2 | 30.8 | 11.9 | 17.8 | 36.4 | 33.3 |
| | IS-PtC | 21.3 | 48.7 | 41.4 | 13.6 | 36.0 | 20.6 | 20.5 | 46.2 | 46.1 |
| **Ours** | UniSPC | 17.3 | 35.5 | 47.5 | 14.9 | 31.5 | 42.3 | 11.7 | 31.4 | 40.5 |
| | RealSPC | 25.5 | 53.5 | 71.8 | 26.7 | 57.2 | 76.9 | 26.4 | 62.6 | 80.9 |

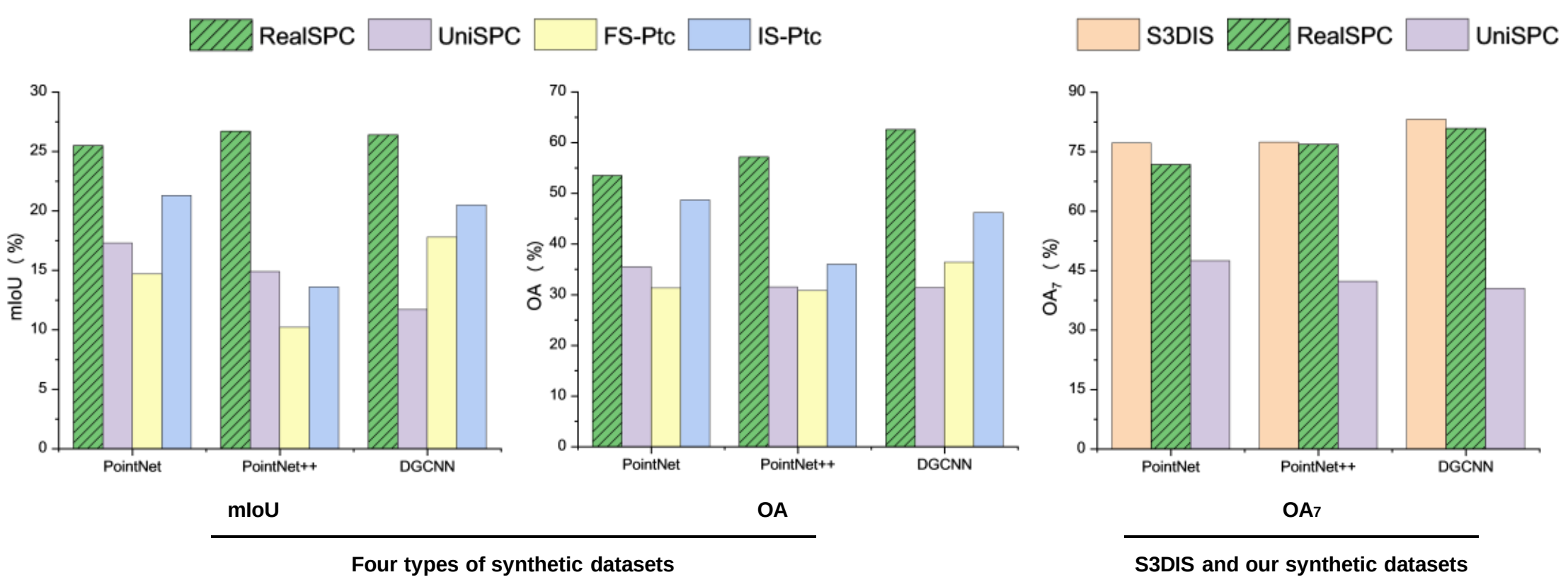

**Fig. 7.** Results of homogenous experiment. Left) RealSPC wins the best mIoU and OA among four types of synthetic datasets. Right) RealSPC achieves comparable $OA_7$ with S3DIS.

### 4.3.2 Augmented training experiment

Experiments in 4.3.1 prove that our synthetic datasets are effective. However, synthetic data cannot replace real data entirely to achieve a comparable performance. Therefore, in this set of experiments, the purely synthetic datasets are augmented with smaller amounts of real data, forming a series of mixed training sets to train models. Two augmented training sets are constructed with synthetic and real point clouds at a ratio of 3:1. Actually, this also meets the assumption that real point clouds are usually difficult to obtain, while theoretically, synthetic point clouds can be infinitely generated from BIM. For comparison, two real training sets with different sizes are also used for experiments to validate the effectiveness of our synthetic data. Four training sets are shown as below.

(1) 44 S3DIS;

(2) 11 S3DIS;

(3) 11 S3DIS & 33 UniSPC;

(4) 11 S3DIS & 33 RealSPC.

In the last three groups of training sets, the 11 real point clouds are the same, and randomly selected from Area 1 of S3DIS. The synthetic point clouds are also randomly selected from their respective datasets. The training and testing methods are the same as those introduced in Section 4.2.

The evaluation metrics for the four experiments are calculated and shown in Table 6. Regarding overall performance, although the augmented training sets don’t achieve the comparable performance as real dataset S3DIS (44), they all surpass the small size dataset S3DIS (11). Taking the mIoU metric as an example, the augmented training sets outperform S3DIS (11) by 3.5 %, 4.3 %, and 3.0 %, respectively. On the one hand, this demonstrates the value of synthetic data for enhancing segmentation performance when adding into a small size of real dataset. On the other hand, this finding also further validates the effectiveness of synthetic point clouds. Furthermore, in terms of $OA_7$, as shown in Fig. 8, the performance of augmented training using RealSPC is 1.4 %, 9.8 %, and 3.2 % higher than that of using UniSPC, respectively. This indicates that when considering the common semantic classes, RealSPC is better than UniSPC, which means that real-colored point clouds have more realistic features than unicolored ones. This undoubtedly reaffirms that color is an important factor affecting model performance.

**Table 6**

Model performance of augmented training experiments (%)

| Training set | PointNet | | | PointNet++ | | | DGCNN | | |
|---|---|---|---|---|---|---|---|---|---|
| | mIoU | OA | $OA_7$ | mIoU | OA | $OA_7$ | mIoU | OA | $OA_7$ |
| S3DIS (44) | 38.1 | 73.5 | 77.2 | 43.3 | 77.4 | 77.4 | 44.3 | 79.5 | 83.2 |
| S3DIS (11) | 31.6 | 67.3 | 74.6 | 33.0 | 65.8 | 64.5 | 30.6 | 68.2 | 67.7 |
| S3DIS & UniSPC | 32.9 | 69.2 | 73.9 | 35.0 | 69.1 | 66.0 | 33.6 | 70.6 | 70.0 |
| S3DIS & RealSPC | 35.1 | 69.0 | 75.3 | 37.3 | 71.4 | 75.2 | 33.2 | 64.9 | 73.2 |

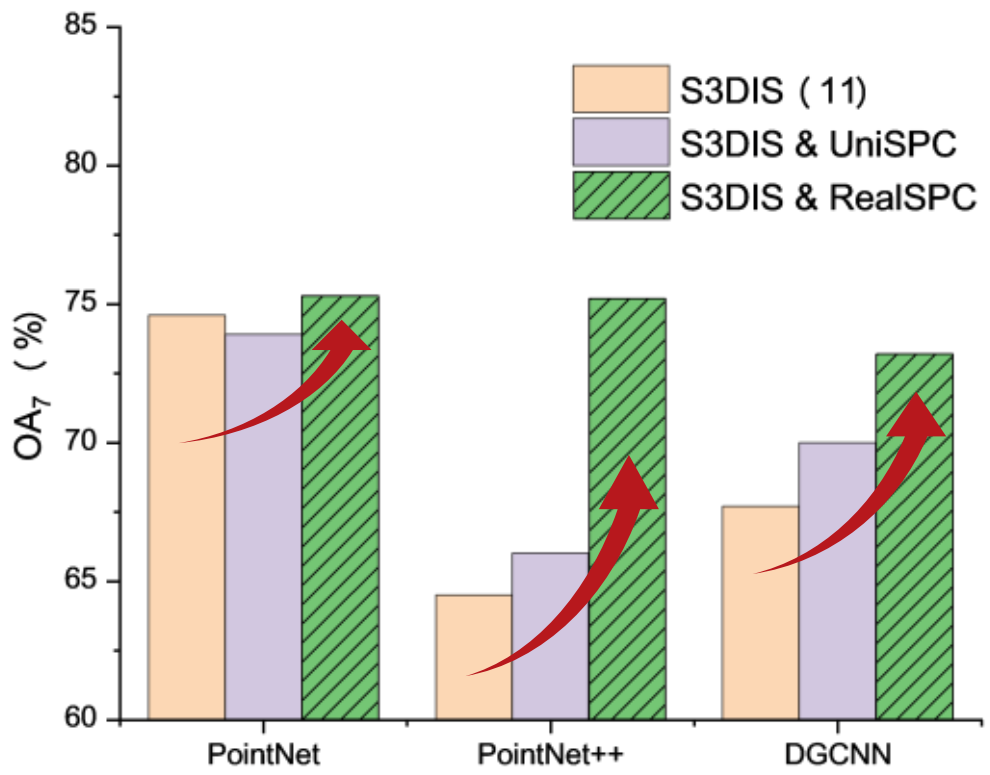

**Fig. 8.** Results of augmented experiment. Adding RealSPC can improve $OA_7$ perfomance of models.

### 4.4 Experiment 2: Effect of SPC proportion on segmentation performance

Experiment 1 demonstrates that SPC color is a significant factor for model training, and RealSPC exhibits a superior effect than UniSPC. Therefore, in mixing experiments, only the impact of different mixing proportions of RealSPC and S3DIS on the segmentation performance is considered. In this study, the mixing proportion is designated as the proportion of synthetic point clouds within the entire training set (the ratio of the number of scenes). In order to gain a more comprehensive result, the mixing proportions of training sets is an arithmetic sequence with a common difference of 5 %, resulting in a total of 21 training sets ranging from 0 to 100 %. 0 % proportion signifies that the training set is entirely composed of S3DIS data, whereas 100 % proportion indicates that the training set is entirely composed of RealSPC. The specific quantities of synthetic and real data in each training set are shown in Fig. 9. Since the total scene number 44 is not a multiple of 20, the specific quantities of point clouds do not constitute a strict arithmetic sequence. Notably, in order to further validate and quantify the value of synthetic data in mixing training, a series of benchmark experiments is conducted in this work. This series of experiments removes the synthetic data from the aforementioned 21 groups of training sets, retaining only the real data from S3DIS. The number of remaining point clouds in training sets is the lower part of Fig. 9. Given that the mixing proportions of 0 % and 100 % just contain homogeneous data, the benchmark experiments for these two groups can be omitted. Ultimately, a total of 21 mixing training sets and 19 benchmark training sets are constructed for the subsequent assessment of segmentation performance. It should be noted that the benchmark experiments are essential in this experiment. This has been elaborately discussed in Section 3.3.1, and the forthcoming results substantiate the significance of benchmark experiments once again. Meanwhile, given the stochastic nature of our random sampling, we train and test each proportion of each model using 3 times of random sampled scenes, and finally take the average metrics as the model performance, to obtain robustness conclusions in different sampled scenes.

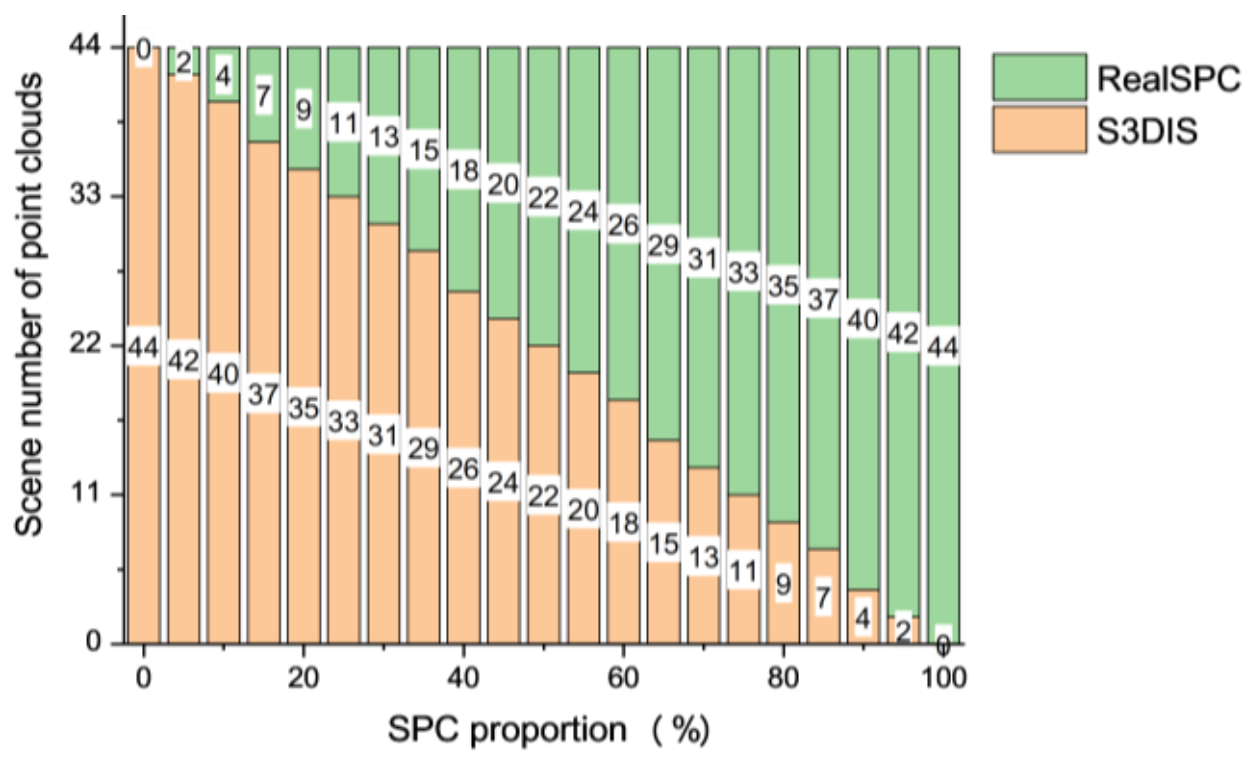

**Fig. 9.** Distribution of point clouds in training sets under different SPC proportions.

**4.4.1 Overall segmentation performance**

Fig. 10 presents the segmentation performance achieved by the 40 groups of training sets on three models. Red lines and blue dashed lines represent the performance of mixing training and the benchmark experiments, respectively. And the bar graph shows the performance deviation between two sets of experiments, where the red bar (positive value) above x-axis indicates better performance of mixing training experiment, while the blue bar (negative value) under x-axis indicates better performance of benchmark experiment. Considering the overall evaluation metrics, when the mixing proportion is less than 70 %, the performance of the mixing training is comparable to that of the benchmark experiments, all maintaining performance without significant degradation as the mixing proportion increases. However, when the mixing proportion exceeds 70 %, due to significant reduction in the size of training sets, the performance of the benchmark experiments begins to decline sharply. For instance, in terms of mIoU, compared to mixing proportion of 70 %, the performance of the benchmark experiment with a 90 % mixing proportion drops by 9.48 %, 10.42 %, and 11.34 %. In contrast, the performance of the mixing training exhibits a smaller decrease when the mixing proportion increases. Similarly, when the mixing proportion increases from 70 % to 90 %, the mIoU decreases by only 1.07 %, 2.36 %, and 7.13 % in mixing training, respectively. As shown in Fig. 10, within the 75 % to 95 % mixing proportion range, all mIoU deviations are positive, indicating that a mixed dataset can achieve better performance than a homogeneous dataset without synthetic point clouds across the three models at higher mixing proportions.

To further quantify the improvement. When the mixing proportion exceeds 70 %, the average improvement of mixing training experiments in evaluation metrics is calculated and shown in Table 7. The results show that, compared with benchmark experiments, the mIoU, OA and $OA_7$ of mixing training are all significantly improved. Especially for the $OA_7$, the improvement achieves 5.88 %, 11.45 % and 14.31 % in the three models, respectively. As the network of models becomes more complex, the improvement in the $OA_7$ metric increases. This implies that synthetic point clouds have a better effect on complex models. Generally speaking, the more complex the network, the stronger its ability to extract features, and the more severe the performance reduction when the training set size is decreased. Therefore, the addition of synthetic point clouds can effectively improve segmentation performance when the mixing proportion is high and the model is complex.

**Table 7**

Average improvement of mixing training compared to benchmark in high mixing proportion (%)

| | **PointNet** | **PointNet++** | **DGCNN** |
|---|---|---|---|
| mIoU | 2.87 | 4.38 | 2.85 |
| OA | 2.21 | 5.80 | 3.81 |
| $OA_7$ | 5.88 | 11.45 | 14.31 |

* High mixing proportion indicates proportions exceed 70 %

Furthermore, taking a conference room point cloud as a test case, the segmentation results of three models with the proportion of 80 % are visualized in Fig. 11. In the error maps (small figures in upper right corner), gray points indicate correct predictions, while red points indicate incorrect predictions. The segmentation results of mixing training are better than those of benchmark experiments, with improvement in OA performance of 2.4 %, 9.0 %, and 10.1 %, respectively. This demonstrates when the mixing proportion is high, the value of synthetic data addition is obvious and significant.

The above results indicate that for the mixing dataset of RealSPC and S3DIS, mixing training has a negligible impact on performance when the mixing proportion is lower than 70 %. However, when the mixing proportion exceeds 70 %, and most real data is replaced by synthetic one, the performance of the mixing training remains relatively stable and shows a significant improvement over the benchmark experiments. Here, we have ample reasons to believe that the benchmark experiments without synthetic point clouds are necessary. Because when the mixing proportion is low, even if synthetic data are removed from the mixed training set, it has no impact on performance, making it difficult to prove the effectiveness of synthetic data. In such cases, the change of model performance in different proportions can be attributed to the change of the scale of real data in the training set but does not indicate the impact of synthetic data. Therefore, the benchmark experiments are necessary for proving the value of synthetic data. In most studies, such benchmark experiments are lacking, making the results one-sided and lacking persuasiveness.

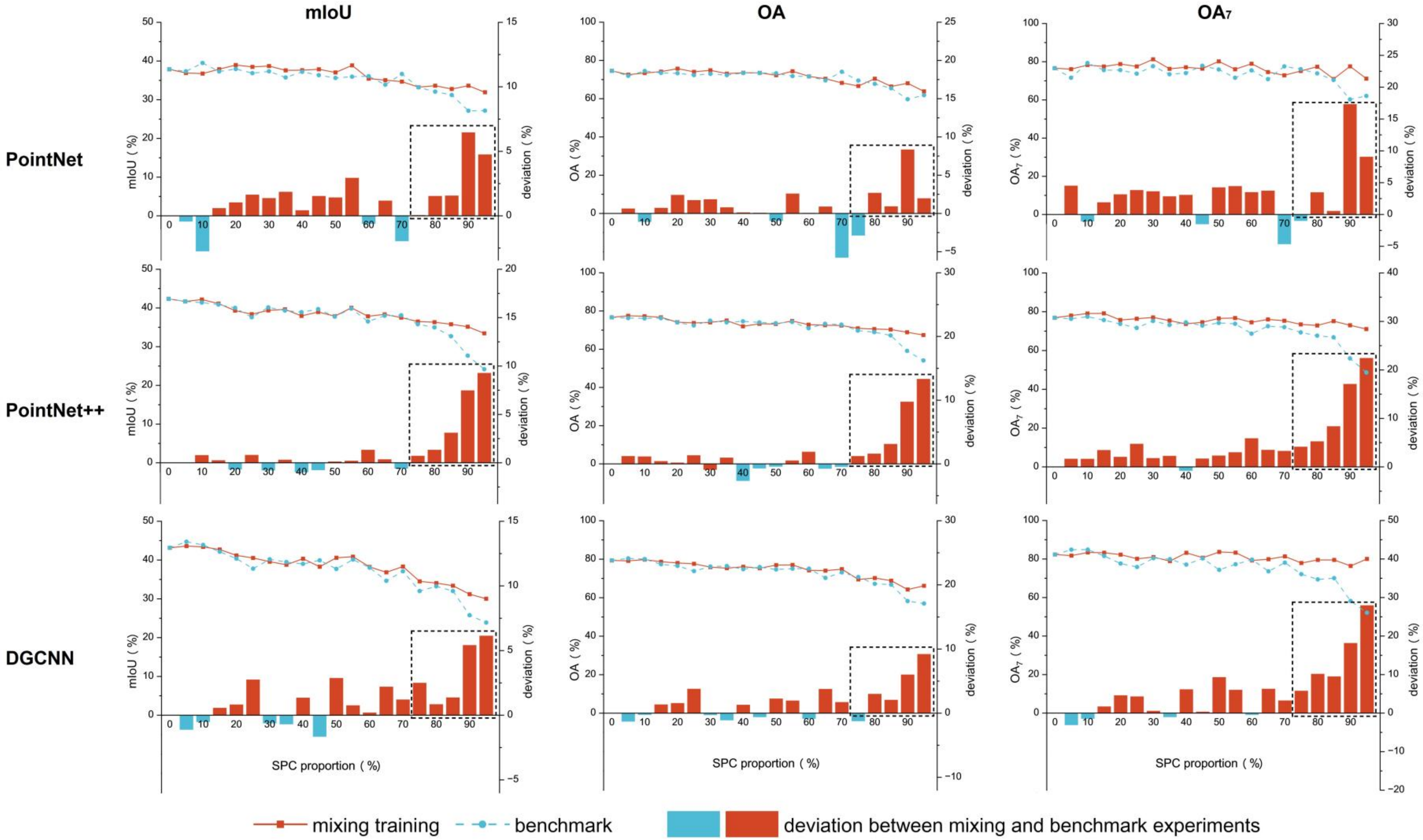

**Fig. 10.** Comparison of mixing proportion experiments between mIoU, OA, $OA_7$ metrics on PointNet, PointNet++, DGCNN models.

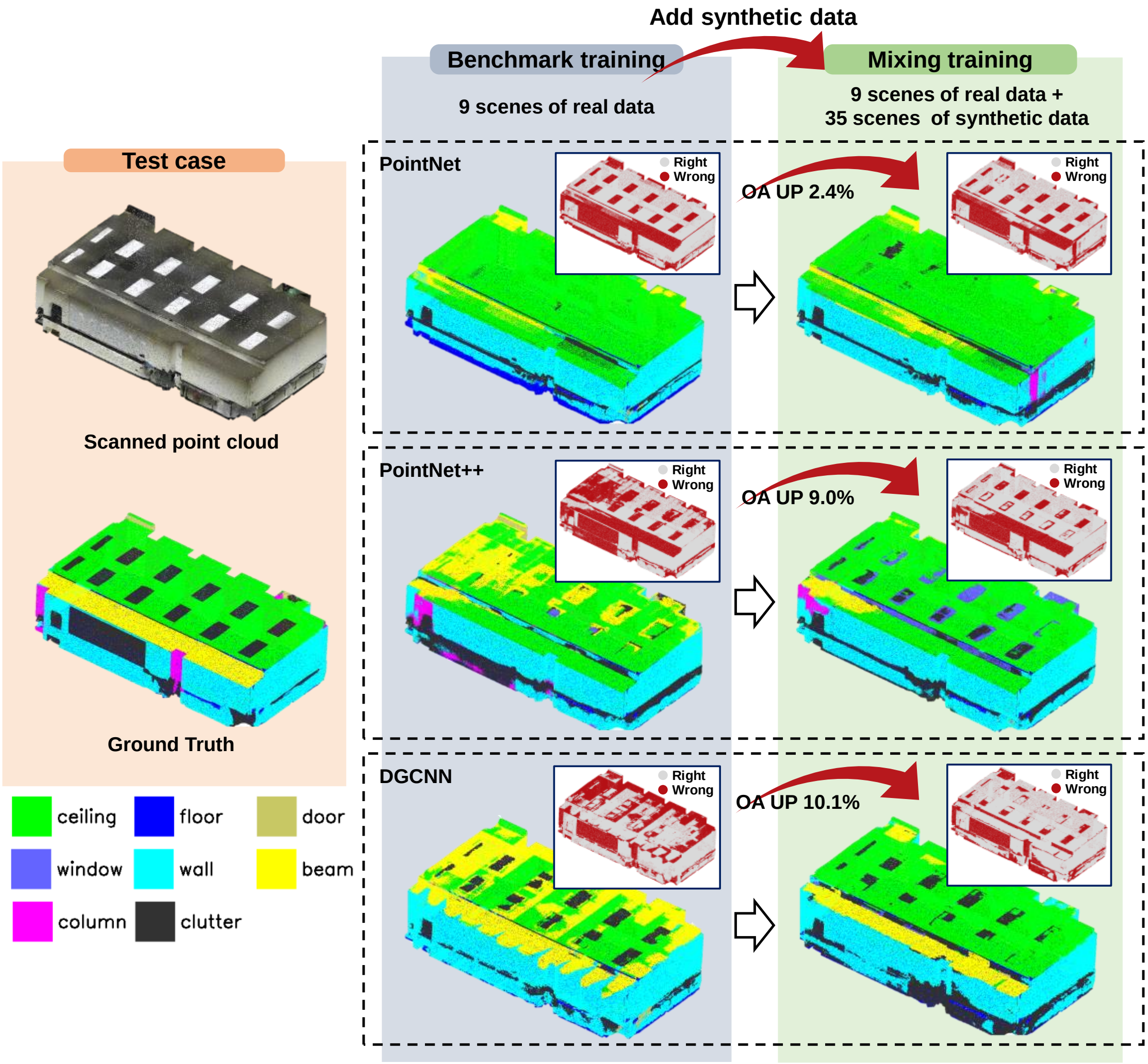

**Fig. 11.** Segmentation visualization under 80 % mixing proportion. Addition of synthetic data obviously improves segmentation effect.

### 4.4.2 Segmentation performance on each class

For further analysis, the accuracy (Acc) deviations in performance of each class between mixing training and benchmark experiment are calculated. And the average Acc deviations (mixing proportions from 5 % to 95 %) are shown in Table 8. This result reveals that the addition of synthetic point clouds has disparate effects on performance across different classes. Notably, the floor class demonstrates the most significant enhancement in all three models, with improvements of 7.9 %, 13.6 %, and 15.5 %, respectively. Meanwhile, the Acc of window, wall and column are also improved overall, although the improvement is not significant on certain models. Furthermore, the impact of adding synthetic data on beam class accuracy is unclear, showing minor effect on PointNet++ and DGCNN, but substantial enhancement on PointNet. However, as expected, obvious negative effects of door and clutter can be found in the result. The former is because the door is open in S3DIS but closed in BIM models, resulting in a gap between real data and synthetic data. While the latter is because there are no clutter points in the synthetic data.

**Table 8**

Impact of adding synthetic point clouds on the accuracy of different classes

| | ceiling | floor | door | window | wall | beam | column | clutter |
|---|---|---|---|---|---|---|---|---|
| PointNet | −3.8 | +7.9 | −2.1 | +0.9 | +6.3 | +7.6 | +0.5 | −6.0 |
| PointNet++ | −1.0 | +13.6 | −6.1 | +10.5 | +5.5 | −0.6 | +3.0 | −8.1 |
| DGCNN | −1.2 | +15.5 | −7.3 | +8.2 | +6.3 | +1.6 | +4.4 | −12.0 |

* Average accuracy deviations when mixing proportions vary from 5 % to 95 %. Positive values indicate adding synthetic data improve the performance of models, while negative values indicate the opposite.

Specifically, Fig. 12 shows the Acc variation trend of each class in mixing training experiments. Notably, clutter class is not considered in this part since clutter points are not included in the synthetic data and have no value for discussion. As shown by the three solid lines in Fig. 12, when the mixing proportion increases, the Acc performance of ceiling, floor and wall classes remain at a high level. This indicates that although real data is constantly replaced by synthetic data, the segmentation ability of these three classes has not decreased, and the Acc performance remains around 80 %, even if only synthetic point clouds are available in the training set (100 % mixing proportion). However, the segmentation performance of other four classes points is not ideal. On the one hand, all three models have limited segmentation ability for these classes. On the other hand, the increase of the mixing proportion cannot guarantee the stable performance of Acc, especially when the mixing proportion is high, the performance often decreases. The above findings indicate that as for those large and flat elements such as walls, ceilings, and floors, synthetic data can replace real data to achieve comparable performance. But the remaining elements (usually small and complex), such as doors, windows, beams, and columns, vary greatly in different buildings. So, there are distinct gaps between the synthetic and real data, making it difficult to reach a consistent agreement in model performance. This is an important finding that can guide us to regenerate synthetic point clouds and reconstruct mixed datasets, for reducing the size of real data while ensuring the performance of models.

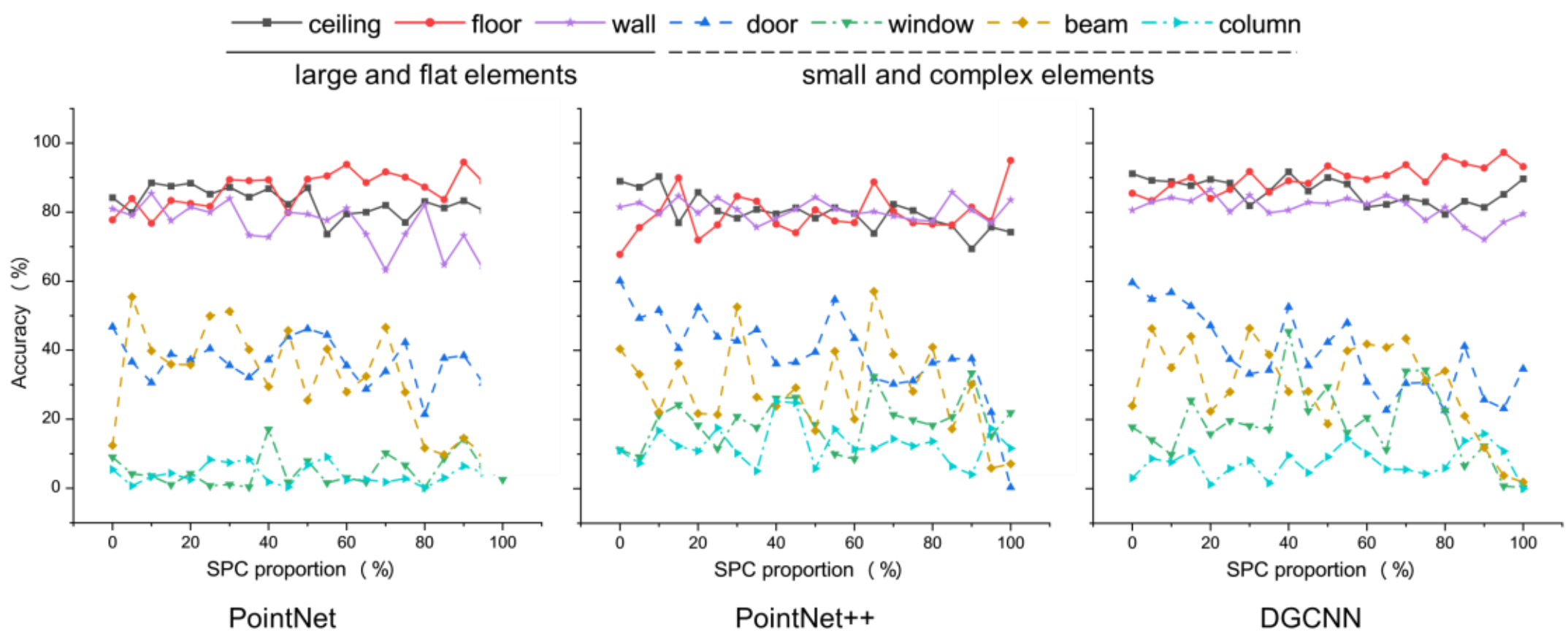

**Fig. 12.** Class accuracy of mixing training under different proportions. Accuracy of large and flat elements maintains at a high level.

## 5 Discussion and limitation

This paper proposes a BIM-based method for synthetic point cloud generation. And on this basis, two types of synthetic datasets with different colors are generated. UniSPC is directly assigned consistent colors from BIM models, while RealSPC is colored through registration with real scanned point clouds. Then two sets of experiments reveal fundamental understandings of the effect of synthetic color and mixing proportion on model performance. Based on the results of experiments, some insights and limitations are discussed as follows.

### 5.1 Synthetic color

Ideally, there should be no discrepancy between the synthetic data and real data, but no existing synthetic data can meet this requirement [41]. This problem also exists in synthetic point clouds, especially in the difficulty of color simulating. The experiments in section 4.3 demonstrate that the color of synthetic point clouds is a critical factor in the performance of segmentation models. If just considering the overall accuracy of seven classes without clutter class ($OA_7$), the homogeneous training set of RealSPC shows comparable performance to the real training set. This finding indicates that our synthetic point clouds with real color have a similar feature with real data. However, there is a problem: the acquisition of real scanned point clouds as same as BIM model is still a challenging. On the one hand, there is a gap between BIM models and real buildings. For example, small components such as fire hydrants, elevators, and furniture are not modeled in BIM, but they exist in the real world. On the other hand, scanning the real buildings is a time-consuming and high-cost task. So, it is still difficult to obtain a large number of synthetic point clouds with real colors through our method. Considering these problems, a method to color synthetic point clouds with a more realistic texture is necessary. But the real colors are influenced by ambient lights or surrounding environments, especially for components such as windows and high reflectivity surfaces. In theory, 3D ray tracing can have a better restoration of real colors, but the rendering process is time-consuming and inefficient. To overcome this issue, transfer learning or reconstruction from 2D images may be effective. The simulation of colors will be a future work to obtain high-quality synthetic point clouds.

### 5.2 Mixing proportion

As expected, mixing proportion is also an important factor for the segmentation performance. In our findings, when the proportion of synthetic data is low, the real data dominates in the training set. The addition of synthetic data has a negligible impact on the model performance. But while the situation reverses, the size of real data is limited, resulting in poor performance of only real training sets. At this point, the addition of synthetic data can improve the performance. In our experiments, when the mixing proportion of synthetic data is over 70 %, the improvement becomes significant. Although these findings are obtained on our datasets and S3DIS, of which the threshold of mixing proportion may differentiate across datasets, it can be foreseen that a high proportion of synthetic data usually leads to better performance. That means even if there is only a small amount of real data, by generating enough amount of synthetic data, the mixing training can achieve a better model performance. Or to put it another way, the size of the dataset is relative to the complexity of the model. In the future, if the network structure of deep learning becomes more and more complex and the volume of trainable parameters becomes larger and larger, the demand of data in model training will also increase. The value of synthetic data will become more prominent. Although high-quality real 3D data is still very scarce, the addition of massive synthetic data holds the promise of achieving superior performance that surpasses the capabilities inherent in real datasets.

### 5.3 Dataset creation

In this paper, the segmentation performance of different classes is various. Some synthetic points of large and flat elements achieve comparable performance with that of real data. Without a doubt, this finding will help us regenerate synthetic point clouds and reconstruct mixed dataset, for achieving a better segmentation performance. For instance, the synthetic points of floors, ceilings, and walls show a similar feature to that of real data, which can replace real data without compromising model performance. This implies that the data collection frequency of these components can be reduced in real data acquisition. And use synthetic point clouds containing a large number of these components to compensate for their absence. Conversely, for the remaining classes, synthetic point clouds cannot accurately represent the features as same as those in real data, requiring more frequent data collection. In theory, we can leverage the advantages of both real data and synthetic data to form a more effective mixed dataset, such as a dataset with synthetic walls, floors, ceilings and real doors, windows etc. Even if we only collect and annotate a small number of real points, mixed datasets can achieve a similar performance as large-scale real datasets. In addition, synthetic point clouds with similar features to real ones provide us with the possibility of training a large and universal 3D segmentation model. As shown in Human3D study of Takmaz et al. [42], pre-training followed by fine-tuning was been demonstrated to improve the performance and generalization of point cloud segmentation of humans. If the synthetic dataset is large enough, pre-training on a large size of synthetic data and fine-tuning on a small size of real data may inspire greater potential for the 3D segmentation model.

## 6. Conclusion

Training semantic segmentation models for point clouds requires a large amount of high-quality data, and synthetic point clouds (SPCs) is an important technique to alleviate the data scarcity. However, how the color and mixing proportion of synthetic point clouds impact the performance of trained segmentation models remains a long-standing question, which, has not been systematically investigated before. Therefore, this paper unveils the performance-improving mechanism of color and mixing proportion of SPC on semantic segmentation with extensive experiments. We first proposed a method for synthetic point cloud generation based on BIM. Considering the range of the scanner, our method can automatically select the scanner stations to simulate the real scanning process. On this basis, we constructed two synthetic datasets: UniSPC (with consistent color from BIM models) and RealSPC (with real color from scanned point clouds). Together with the S3DIS dataset, comprehensive experiments on PointNet, PointNet++ and DGCNN models were conducted to investigate the impact of synthetic color and mixing proportion on segmentation performance. And in this work, a series of benchmark experiments and an evaluation metric ($OA_7$) were proposed for reliable results. First, in homogenous experiments, our RealSPC achieved a comparable segmentation performance on $OA_7$ to that of real data from S3DIS, while also showed improvement of 8.2 % + in mIoU and 18.0 % + in OA over UniSPC, respectively. Second, in augmented training experiments, the augmented training sets outperformed the real training sets without synthetic data by 3.5 %, 4.3 %, and 3.0 % on mIoU metric, respectively. Third, in the experiments of mixing proportions, when the mixing proportion was lower than 70 %, there was a negligible impact of synthetic point clouds on performance compared to the benchmark. However, when the mixing proportion exceeded 70 %, the mixing training achieved average improvements of 5.9 %, 11.5 % and 14.3 % in $OA_7$ for the three models, respectively. Furthermore, the segmentation accuracy of different classes showed different patterns, in which synthetic points of large and flat elements, such as floors, walls and ceilings, achieved similar performance with the real points, while others were ambiguous or weaken. These key findings can be summarized as follow: (1) the synthetic point clouds generated by this study is effective; (2) synthetic color is a critical factor on the segmentation performance, and real color is better than consistent color of BIM models; (3) mixing proportion is also important for model training, showing when the real data is limited and the mixing proportion is high, the addition of synthetic data can significantly improve the segmentation performance; (4) for large and flat elements in built environment, synthetic data can even replace real ones without compromising model performance.

This paper provides a feasible method to generate synthetic point clouds, which can alleviate the shortage of 3D data on some degrees. The roles of synthetic colors and mixing proportions are investigated by a series of comprehensive experiments. Furthermore, the value of synthetic point clouds is quantified. The underlying performance-improving mechanism of these two factors on semantic segmentation are revealed in this paper, which contributes to the body of knowledge and application of synthetic point clouds. Especially for the field of building and construction, the proposed synthetic point clouds generation method can improve the performance of classical segmentation task and provide possibilities for more engineering applications.

Further research is required to expand the classes of synthetic point clouds and construct a large size of synthetic dataset for fully exploring the value of our synthetic data. More real datasets will be used for evaluation to obtain more general conclusions and guide 3D perception tasks in more and larger scenes. Meanwhile, more advanced semantic segmentation models will be considered to investigate the potential of synthetic point clouds under different network architectures (the latest results can be found in https://github.com/smartaec/Synthetic-Point-Clouds-

for-semantic-segmentation). Furthermore, some potential training methods will also be explored in the future, such as strategy that pre-train in synthetic dataset and fine-tune in real dataset, for fully utilizing the value of synthetic point clouds.

**Acknowledgments**

This research was conducted with the supports of the "National Key R&D Program of China" (Project No. 2023YFC3805800) and the project "Research on Key Technologies for Mechanized Construction of Substation Civil Engineering" of "Technology project funding from State Grid Corporation of China" (Project No. 5200-202311481A-3-2-ZN).

**Reference**

[1] J.J. Chen, Q. Su, Y.B. Niu, Z.Y. Zhang, J.H. Liu, A handheld LiDAR-based semantic automatic segmentation method for complex railroad line model reconstruction, Remote Sensing 15 (18) (2023). https://doi.org/10.3390/rs15184504

[2] K. Han, J. Degol, M. Golparvar-Fard, Geometry- and appearance-based reasoning of construction progress monitoring, J. Constr. Eng. Manag. 144 (2) (2018). https://doi.org/10.1061/(Asce)Co.1943-7862.0001428

[3] N. Bolourian, M. Nasrollahi, F. Bahreini, A. Hammad, Point cloud-based concrete surface defect semantic segmentation, Journal of J. Comput. Civ. Eng. 37 (2) (2023). https://doi.org/10.1061/Jccee5.Cpeng-5009

[4] C.R. Qi, H. Su, K.C. Mo, L.J. Guibas, PointNet: Deep Learning on Point Sets for 3D Classification and Segmentation, 30th Ieee Conference on Computer Vision and Pattern Recognition (Cvpr 2017) (2017) 77-85. http://dx.doi.org/10.1109/CVPR.2017.16

[5] Y. Wang, Y.B. Sun, Z.W. Liu, S.E. Sarma, M.M. Bronstein, J.M. Solomon, Dynamic graph CNN for learning on point clouds, ACM Trans. Graph. 38 (5) (2019) 1-12, https://doi.org/10.1145/3326362

[6] S. Savarese, 3D semantic parsing of large-scale indoor spaces, in: Proceedings of the IEEE Conference on Computer Vision and Pattern Recognition, 2016, pp. 1534–1543, https://doi.org/10.1109/CVPR.2016.170

[7] R. Duan, H. Deng, M. Tian, Y.C. Deng, J.R. Lin, SODA: a large-scale open site object detection dataset for deep learning in construction, Automation in Construction 142 (2022), p. 104499. https://doi.org/10.1016/j.autcon.2022.104499

[8] M. Carós, A. Just, S. Seguí, J. Vitrià, Self-Supervised Pre-Training Boosts Semantic Scene Segmentation on LiDAR data, 2023 18th International Conference on Machine Vision and Applications, Mva (2023). https://doi.org/10.23919/MVA57639.2023.10216191

[9] Y.C. Song, Z.X. Sun, Y.J. Wu, Y.H. Sun, S.T. Luo, Q. Li, Learning Semantic Segmentation on Unlabeled Real-World Indoor Point Clouds via Synthetic Data, 2022 26th International Conference on Pattern Recognition (Icpr) (2022) 3750-3756. https://doi.org/10.1109/ICPR56361.2022.9956612

[10] M.L. Wei, Few-shot 3D point cloud semantic segmentation with prototype alignment, proceedings of 2023 8th international conference on machine learning technologies, Icmlt 2023 (2023) 195-200. https://doi.org/10.1145/3589883.3589913
[11] G. Paulin, M. Ivasic-Kos, Review and analysis of synthetic dataset generation methods and techniques for application in computer vision, Artif. Intell. Rev. 56 (9) (2023) 9221-9265. https://doi.org/10.1007/s10462-022-10358-3
[12] M. Afzal, R.Y.M. Li, M. Shoaib, M.F. Ayyub, L.C. Tagliabue, M. Bilal, H. Ghafoor, O. Manta, Delving into the Digital Twin Developments and Applications in the Construction Industry: A PRISMA Approach, Sustainability 15 (23) (2023). https://doi.org/10.3390/su152316436
[13] J.W. Ma, T. Czerniawski, F. Leite, Semantic segmentation of point clouds of building interiors with deep learning: augmenting training datasets with synthetic BIM-based point clouds, Automation in Construction 113 (2020). https://doi.org/10.1016/j.autcon.2020.103144
[14] S. Tang, H. Huang, Y. Zhang, M. Yao, X. Li, L. Xie, W. Wang, Skeleton-guided generation of synthetic noisy point clouds from as-built BIM to improve indoor scene understanding, Automation in Construction 156 (2023). https://doi.org/10.1016/j.autcon.2023.105076
[15] F. Noichl, F.C. Collins, A. Braun, A. Borrmann, Enhancing point cloud semantic segmentation in the data-scarce domain of industrial plants through synthetic data, Comput. Aided Civ. Inf. Eng. (2024). https://doi.org/10.1111/mice.13153
[16] Y.L. Guo, H.Y. Wang, Q.Y. Hu, H. Liu, L. Liu, M. Bennamoun, Deep learning for 3D point clouds: a survey, IEEE Trans. Pattern Anal. Mach. Intell. 43 (12) (2021) 4338-4364. https://arxiv.org/pdf/1912.12033.pdf
[17] D.P. Singh, M. Yadav, Deep learning-based semantic segmentation of three-dimensional point cloud: a comprehensive review, Int. J. Remote Sens. 45 (2) (2024) 532-586. https://doi.org/10.1080/01431161.2023.2297177
[18] Z.J. Liu, H.T. Tang, Y.J. Lin, S. Han, Point-Voxel CNN for Efficient 3D Deep Learning, Advances in Neural Information Processing Systems 32 (Nips 2019) 32 (2019). https://arxiv.org/abs/1907.03739
[19] C.R. Qi, L. Yi, H. Su, L.J. Guibas, PointNet++: Deep Hierarchical Feature Learning on Point Sets in a Metric Space, Advances in Neural Information Processing Systems 30 (Nips 2017) 30 (2017). https://proceedings.neurips.cc/paper_files/paper/2017/file/d8bf84be3800d12f74d8b05e9b89836f-Paper.pdf
[20] Q.Y. Hu, B. Yang, L.H. Xie, S. Rosa, Y.L. Guo, Z.H. Wang, N. Trigoni, A. Markham, RandLA-net: efficient semantic segmentation of large-scale point clouds, Proceedings of the IEEE/CVF conference on computer vision and pattern recognition (2020) 11108-11117. http://openaccess.thecvf.com/content_CVPR_2020/papers/Hu_RandLA-Net_Efficient_Semantic_Segmentation_of_Large-Scale_Point_Clouds_CVPR_2020_paper.pdf

[21] C. Choy, J. Gwak, S. Savarese, 4D Spatio-Temporal ConvNets: Minkowski Convolutional Neural Networks, 2019 Ieee/Cvf Conference on Computer Vision and Pattern Recognition (Cvpr 2019) (2019) 3070-3079. https://doi.org/10.1109/Cvpr.2019.00319

[22] M.H. Guo, J.X. Cai, Z.N. Liu, T.J. Mu, R.R. Martin, S.M. Hu, PCT: point cloud transformer, Comput. Vis. Media 7 (2) (2021) 187-199. https://doi.org/10.1007/s41095-021-0229-5

[23] H.S. Zhao, L. Jiang, J.Y. Jia, P. Torr, V. Koltun, Point Transformer, 2021 Ieee/Cvf International Conference on Computer Vision (Iccv 2021) (2021) 16239-16248. https://arxiv.org/pdf/2210.05666

[24] X.Y. Wu, Y.X. Lao, L. Jiang, X.H. Liu, H.S. Zhao, Point Transformer V2: Grouped Vector Attention and Partition-Based Pooling, Advances in Neural Information Processing Systems 35 (Neurips 2022) (2022). https://doi.org/10.1109/Iccv48922.2021.01595

[25] X.Y. Wu, J. Li, P.S. Wang, Z.J. Liu, X.H. Liu, Q. Yu, W.L. Ouyang, T. He, H.S. Zhao, Point Transformer V3: simpler, faster, stronger, 2024 IEEE/CVF Conference on Computer Vision and Pattern Recognition (CVPR), 2024, 4840-4851. https://doi.org/10.1109/CVPR52733.2024.00463

[26] A. Dai, A.X. Chang, M. Savva, M. Halber, T. Funkhouser, M. Niessner, ScanNet: Richly-annotated 3D Reconstructions of Indoor Scenes, 30th Ieee Conference on Computer Vision and Pattern Recognition (Cvpr 2017) (2017) 2432-2443. https://doi.org/10.1109/Cvpr.2017.261

[27] J. Tobin, R. Fong, A. Ray, J. Schneider, W. Zaremba, P. Abbeel, Domain Randomization for Transferring Deep Neural Networks from Simulation to the Real World, 2017 Ieee/Rsj International Conference on Intelligent Robots and Systems (Iros) (2017) 23-30. https://doi.org/10.1109/IROS.2017.8202133

[28] A. Larumbe, M. Ariz, J.J. Bengoechea, R. Segura, R. Cabeza, A. Villanueva, Improved Strategies for HPE Employing Learning-by-Synthesis Approaches, 2017 Ieee International Conference on Computer Vision Workshops (Iccvw 2017) (2017) https://doi.org/10.1109/ICCVW.2017.182

[29] G. Ros, L. Sellart, J. Materzynska, D. Vazquez, A.M. Lopez, The SYNTHIA Dataset: A Large Collection of Synthetic Images for Semantic Segmentation of Urban Scenes, 2016 Ieee Conference on Computer Vision and Pattern Recognition (Cvpr) (2016) 3234-3243. https://doi.org/10.1109/CVPR.2016.352

[30] Q. Wang, J.Y. Gao, W. Lin, Y. Yuan, Learning from Synthetic Data for Crowd Counting in the Wild, 2019 Ieee/Cvf Conference on Computer Vision and Pattern Recognition (Cvpr 2019) (2019) 8190-8199. https://doi.org/10.1109/Cvpr.2019.00839

[31] R. Cazorla, L. Poinel, P. Papadakis, C. Buche, Reducing domain shift in synthetic data augmentation for semantic segmentation of 3D point clouds, 2022 IEEE International Conference on Systems, Man, and Cybernetics (SMC), Prague, Czech Republic, 2022, pp. 1198-1205. https://doi.org/10.1109/SMC53654.2022.9945480

[32] R.M. Zhai, J.G. Zou, Y.F. He, L.Y. Meng, BIM-driven data augmentation method for semantic segmentation in superpoint-based deep learning network, Automation in Construction 140 (2022). https://doi.org/10.1016/j.autcon.2022.104373

[33] J.W. Ma., B. Han., F. Leite., An Automated Framework for Generating Synthetic Point Clouds from as-Built BIM with Semantic Annotation for Scan-to-BIM, 2021 Winter Simulation Conference (WSC), Phoenix, AZ, USA, 2021, pp. 1-10. https://doi.org/10.1109/WSC52266.2021.9715301

[34] H.X. Zhang, Z.B. Zou, Quality assurance for building components through point cloud segmentation leveraging synthetic data, Autom. Constr. 155 (2023). https://doi.org/10.1016/j.autcon.2023.105045

[35] buildingSMART International Ltd., IfcSpace, Retrieved on 05/12/2024 from, 2017. https://standards.buildingsmart.org/IFC/RELEASE/IFC4_1/FINAL/HTML/schema/ifcproductextension/lexical/ifcspace.htm

[36] Q.-Y. Zhou, J. Park, V. Koltun, Open3D: A modern library for 3D data processing, 2018, arXiv:1801.09847. https://doi.org/10.48550/arXiv.1801.09847

[37] F. Bahreini, A. Hammad, Dynamic graph CNN based semantic segmentation of concrete defects and as-inspected modeling, Automation in Construction 159 (2024). https://doi.org/10.1016/j.autcon.2024.105282

[38] H. Hu, Q. Tan, R.H. Kang, Y.L. Wu, H. Liu, B.G. Wang, Building extraction from oblique photogrammetry point clouds based on PointNet plus plus with attention mechanism, Photogrammetric Record 39 (185) (2024) 141-156. https://doi.org/10.1111/phor.12476

[39] Z. Xiong, T. Wang, Research on BIM reconstruction method using semantic segmentation point cloud data based on PointNet, IOP Conference Series: Earth and Environmental Science 719 (2021) 022042. https://doi.org/10.1088/1755-1315/719/2/022042

[40] C.C.Y. Zhou, Y.Q. Dong, M.L. Hou, Y.H. Ji, C.H. Wen, MP-DGCNN for the semantic segmentation of Chinese ancient building point clouds, Heritage Science 12 (1) (2024). https://doi.org/10.1186/s40494-024-01289-z

[41] L. Xu, H. Liu, B. Xiao, X. Luo, Z. Zhu, Synthetic Simulated Data for Construction Automation: A Review, Construction Research Congress, American Society of Civil Engineers, Des Moines, Iowa, 2024, pp. 527-536. https://doi.org/10.1061/9780784485262.054

[42] A. Takmaz, J. Schult, I. Kaftan, M. Akçay, B. Leibe, R. Sumner, F. Engelmann, S.Y. Tang, 3D Segmentation of Humans in Point Clouds with Synthetic Data, 2023 Ieee/Cvf International Conference on Computer Vision, Iccv (2023) 1292-1304. https://doi.org/10.1109/Iccv51070.2023.00125

## Appendix A Loss curves and mIoU curves of model training

See Fig. A.

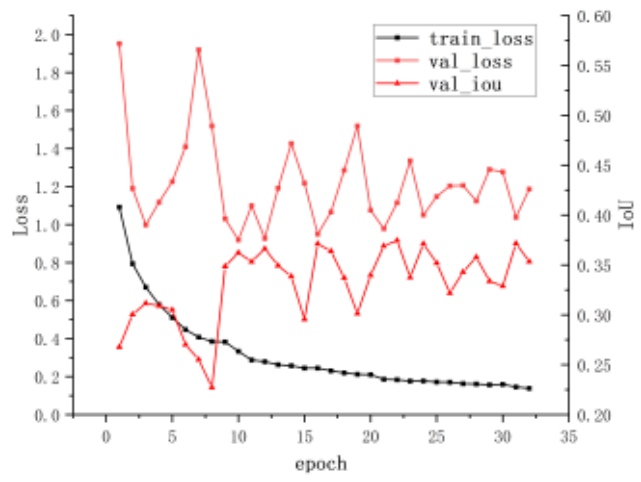

PointNet, mixing experiment, 5%

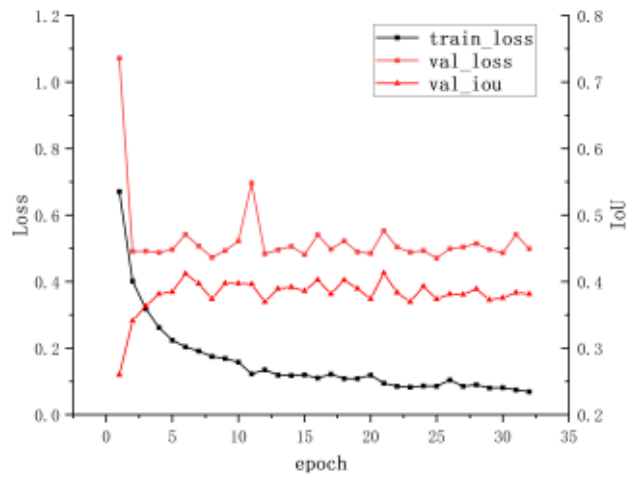

PointNet++, mixing experiment, 5%

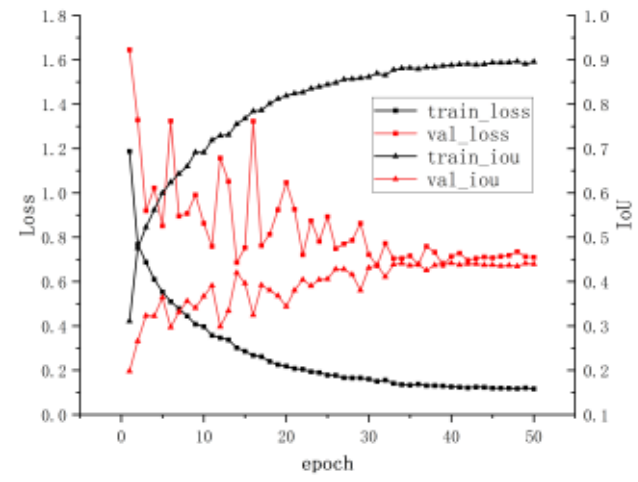

DGCNN, mixing experiment, 5%

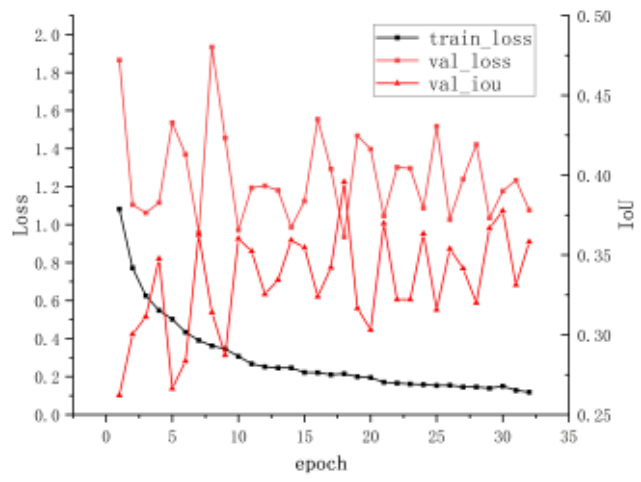

PointNet, benchmark, 5%

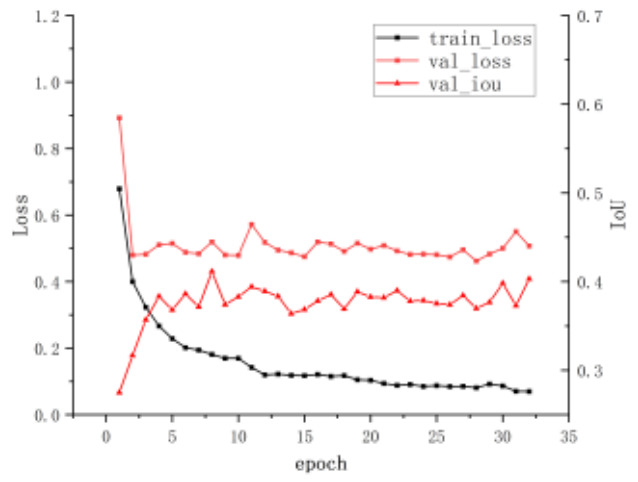

PointNet++, benchmarkt, 5%

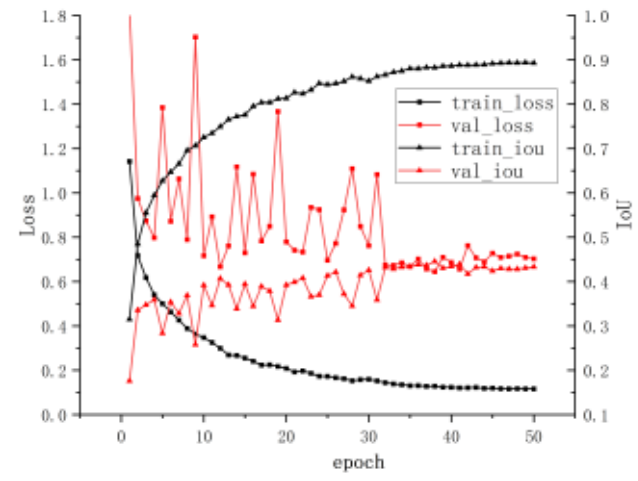

DGCNN, benchmark, 5%

**Fig. A** Loss curves and mIoU curves: all models are sufficiently trained